\documentclass{article}

\usepackage[main, final]{neurips_2026}

\usepackage{graphicx}
\usepackage{pifont}
\usepackage{bbding}
\usepackage{colortbl}
\usepackage{amsmath}

\usepackage{multirow}
\usepackage{adjustbox}

\usepackage{fmtcount}

\usepackage{fontawesome5}

\usepackage[utf8]{inputenc} 
\usepackage[T1]{fontenc}    
\usepackage{hyperref}       
\usepackage{url}            
\usepackage{booktabs}       
\usepackage{amsfonts}       
\usepackage{nicefrac}       
\usepackage{microtype}      
\usepackage{xcolor}         

\definecolor{gray60}{gray}{.60}
\definecolor{gray90}{gray}{.90}
\definecolor{gray92}{gray}{.92}
\definecolor{gray94}{gray}{.94}
\definecolor{gray96}{gray}{.96}
\newcommand{\grow}[1]{\rowcolor{gray94}{#1}} 

\definecolor{vtpblue}{RGB}{218, 232, 242}    
\definecolor{vtprose}{RGB}{242, 220, 219}    

\title{Virtual Task Prompting for Multi-Task Scene Understanding}

\author{%
  Zedong Wang$^{*}$\quad Yucheng Wang\thanks{Equal contribution. \quad $^{\dagger}$Corresponding author.} \quad Dan Xu$^{\dagger}$ \\[0.3em]
  The Hong Kong University of Science and Technology\\[0.2em]
  \texttt{\{zwangmw, ywangls\}@connect.ust.hk} \quad \texttt{danxu@cse.ust.hk}\\
}

\begin{document}

\maketitle

\begin{abstract}
  Multi-task scene understanding models typically rely on modified vision encoders or complex decoders for cross-task interaction, which limits their flexibility and compatibility with off-the-shelf vision foundation models (VFMs). We introduce \underline{\textbf{V}}irtual \underline{\textbf{T}}ask \underline{\textbf{P}}rompting (VTP), which frames multi-task prompting as retrieving task-adaptive compositions from a shared virtual state, where both the state and routing process are optimized in an end-to-end manner. Concretely, VTP maintains an expanded set of virtual tasks alongside the backbone. At each layer, learnable routing matrices read task-specific compositions as input to interact with image tokens and write the processed results back, forming explicit pathways that capture the intricacies among tasks outside the encoder. Since VTP keeps the backbone clean, it integrates seamlessly with pretrained VFMs, advancing the state of the art on Pascal-Context and NYUD-v2. Stress test on Taskonomy with 13 heterogeneous tasks further shows VTP's robustness in extreme settings. Ablations and analysis indicate that VTP's gains stem from adaptive routing rather than increased capacity. The project page is available at \url{https://jacky1128.github.io/VTP/}.
\end{abstract}    
\section{Introduction}
\label{sec:intro}

Multi-Task Learning (MTL)\cite{caruana1997MultiTask} trains a single model simultaneously over multiple correlated tasks to learn robust, generalizable representations~\cite{icml24PlatonicRep}. It has driven rapid progress in computer vision~\cite{doersch2017MTLVision, kirillov2023SAM, wang2026careedit}, particularly in scene understanding for autonomous driving~\cite{chowdhuri2019multinet, ishihara2021multi} and embodied AI~\cite{chaplot2019embodied, hu2021unit}. As Transformers~\cite{vaswani2017attention, dosovitskiy2020image} become the de facto choice for vision tasks, multi-task scene understanding models have followed suit, which typically fall into two groups: (i) Hard-parameter sharing (HPS) employs a shared encoder for task-agnostic representations, followed by specifically-designed decoders, \textit{e.g.}, cross-attention~\cite{ye2022invpt, ye2024invplpl}, state space models~\cite{lin2024mtmamba}, or mixture-of-experts~\cite{ye2023taskexpert}, for decoding-stage interaction. (ii) Soft-parameter sharing (SPS)~\cite{sun2020AdaShare} instead maintains multiple parallel backbones, enabling cross-task message passing throughout the entire network architecture.

TaskPrompter~\cite{ye2023taskprompter} sets an alternative to both. It appends learnable task vectors for each task that are concatenated with input image tokens to interact directly through Vision Transformers (ViTs)~\cite{dosovitskiy2020image, touvron2021DEiT}, which combines the parameter efficiency of HPS with the input-to-output inter-task communication in SPS, while replacing complex decoders with lightweight prompt-guided prediction heads. It shows the potential to capture both task-specific and agnostic information at once in a streamlined model.

\begin{figure}[h]
    \centering
    \includegraphics[width=\textwidth]{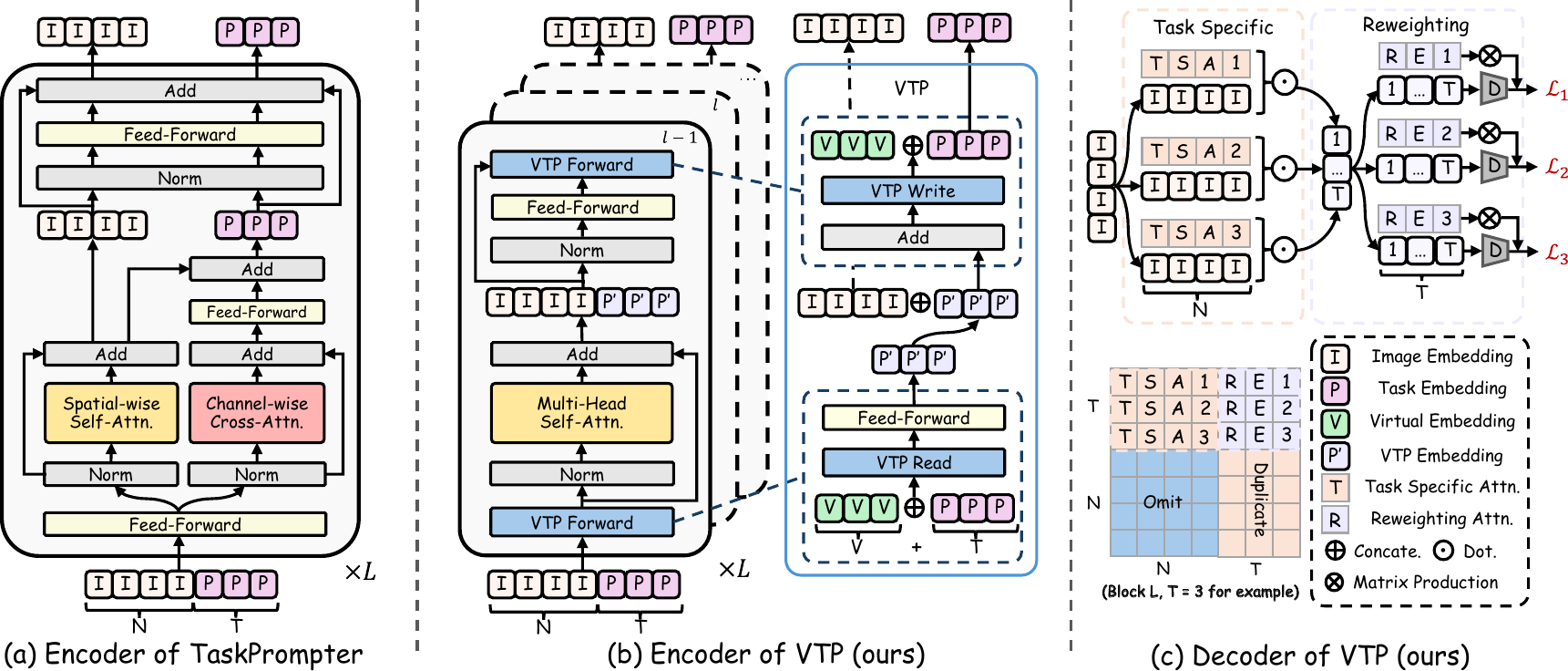}
    \vspace{-1.30em}
        \caption{\textbf{Overview of Virtual Task Prompting (VTP).} (a)~TaskPrompter~\cite{ye2023taskprompter} modifies ViT encoder to compensate for the limited capacity. (b)~VTP instead maintains $T{+}V$ virtual states alongside the encoder, where the model read them to $T$ task vectors to interact with image tokens, and then write the results back to $T{+}V$ bypass, enriching task representations while maintaining a vanilla ViT block. (c)~VTP keeps the prompt-guided decoding pipeline but unifies the spatial-channel parts into one.}
        \vspace{-1.20em}
    \label{fig:VTP}
\end{figure}

However, multi-task prompting has yet to deliver on its promise. As shown in Figure~\ref{fig:VTP}(a), in order to compensate for its simplified architecture, TaskPrompter~\cite{ye2023taskprompter} partitions task vectors into spatial and channel components for increased capacity, which induces inevitable modifications to the vision encoder (\textit{e.g.}, extra channel attention, channel-wise sliding window) and complicates the decoding with dual-branch processing. More importantly, since most existing vision foundation models (VFMs) standardize the macro design of ViTs for scaling up and open-source their corresponding weights to empower the community~\cite{Oquab2023DINOv2, simeoni2025dinov3}, such heuristic re-engineering of vision encoder limits the method's flexibility, making it difficult to leverage the power of off-the-shelf VFMs (\textit{e.g.}, weight mismatches). As a result, the MTL community remains largely restricted to the ImageNet priors~\cite{dosovitskiy2020image, ye2022invpt, ye2023taskexpert}, which hinders both its research progress and real-world applications under limited jointly annotated data.

To push beyond the status quo, we ground this work in a specific question: \textit{how to achieve effective task prompting with sufficient capacity while maintaining a vanilla encoder backbone?} 
The prompt setup and learning process first caught our attention. Previous studies assign each task a single fixed vector updated via backpropagation through the entire model. Intuitively, however, correlated tasks may benefit from sharing a common vector, whereas a complex task might require complementary knowledge from distinct vectors to capture its intricacies. A natural remedy is to increase the vector count for capacity (\textit{e.g.}, pairing two per task), yet doubling or quadrupling the vectors yields negligible or even negative gains as shown in Table~\ref{tab:ablation}. To dig deep, further analysis in Figure~\ref{fig:umap}(a) reveals that the task vectors undergo severe mode collapse across layers, converging to near-identical correlation patterns that might mislead the training into subpar representations, resulting in inefficiency and non-generalizability. Thus, we argue that if we can derive an adaptive prompting method that learns to encode the most expressive patterns for each task to interact with image tokens at different layers.

As such, we propose Virtual Task Prompting (VTP), which achieves effective scene understanding for up to 13 tasks without modifying the encoder's internal topology. Inspired by hyper-connections~\cite{zhu2024hyperconnections, xie2025mhc}, we cast multi-task prompting as retrieving specific task vector compositions from a larger virtual state, where the vectors can be diverse or partially overlapping, and both the state and the retrieving process are end-to-end optimizable. Concretely, VTP maintains a set of states of \textit{virtual task} alongside the encoder, expanding the task embedding space from $T$ to $T{+}V$. At each ViT block, learnable routing matrices adaptively reads $T$ task compositions from the $T{+}V$ virtual state for interaction with image tokens, then writes the results back, providing explicit learning pathways that encourages task specialization and cross-task sharing to emerge jointly outside encoder (Sec.~\ref{subsubsec:vtp_encoding}). Each task's representation is thus no longer a fixed vector like static weights but a learned composition drawn from the richer virtual state, where different tasks may share overlappings or access distinct ones. Since VTP operates outside the encoder backbone, it enables seamless integration with diverse VFMs.

We validate VTP on three axes. (i)~Standard benchmarks (Sec.~\ref{subsec:exp_scene_bench}): On Pascal-Context~\cite{chen2014PascalContext} and NYUD-v2~\cite{silberman2012NYUv2}, our VTP shows competitive results ($+2.77\%$ and $+1.95\%$). Integrating VFMs~\cite{simeoni2025dinov3, wang2026vit5} further yields a leap ($+5.15\%$ and $+3.80\%$). (ii)~Stress Test (Sec.~\ref{subsec:exp_tiny_taskonomy}): On Taskonomy~\cite{zamir2018taskonomy} with heterogeneous 13 tasks, VTP offers consistent gains (up to $+9.58\%$), showcasing robustness as the task count grows. (iii)~\textit{Analysis}: Naively scaling vectors yields negligible gains, indicating that VTP's gains stem not only from its added capacity. It also shows lower parameter count (342M vs.\ 401M), higher throughput (Table~\ref{tab:ablation}), and more discriminative task patterns (Figure~\ref{fig:umap}) than the prior baseline.

Our contributions can be summarized as follows:
\begin{itemize}
\item We introduce Virtual Task Prompting (VTP), a prompting-based architecture that eliminates the need for inner modifications to the vision encoder for multi-task scene understanding.
\item By keeping the encoder backbone clean, VTP integrates seamlessly with different off-the-shelf VFMs (\textit{e.g.}, DINOv3~\cite{simeoni2025dinov3} and ViT-5~\cite{wang2026vit5}), substantially advancing the state-of-the-art on standard Pascal-Context~\cite{chen2014PascalContext} and NYUD-v2~\cite{silberman2012NYUv2} without additional modifications.
\item We stress-test VTP on Taskonomy~\cite{zamir2018taskonomy} with 13 heterogeneous tasks covering both classification and dense regression, where it offers consistent gains. Ablations show that VTP's effectiveness stems from its adaptive read-write routing, not merely increased capacity.
\end{itemize}
\section{Related Work}
\label{sec:related}
Multi-task learning seeks to build shared representations that capture both task-specific and cross-task structure. Existing studies pursue this typically from two orthogonal paths: \textit{training algorithms} that re-balance objectives or gradients, and \textit{model architectures} that operate how task representations are built and interact. VTP falls into the latter, enriching task representations through adaptive prompting. 

\vspace{-0.60em}
\paragraph{Multi-Task Optimization.}
Loss scaling balances task objectives by adjusting their weights, progressing from uniform weighting~\cite{zhang2021survey} and homoscedastic uncertainty~\cite{kendall2018UW} to strategies driven by loss trajectories~\cite{liu2019DWA}, geometric means~\cite{chennupati2019GLS}, equal-rate reduction~\cite{liu2024FAMO}, and progress alignment~\cite{shen2024GO4Align}. Gradient manipulation instead modifies task gradients w.r.t.\ shared parameters, encompassing magnitude balancing~\cite{chen2018GradNorm}, conflict-resolving projection~\cite{yu2020PCGrad}, multi-objective Pareto optimization~\cite{desideri12MGDA} with extensions to worst-case improvement~\cite{liu2021CAGrad} and Nash bargaining~\cite{navon2022NashMTL}, principal component alignment~\cite{senushkin2023AlignedMTL}, and utility maximization~\cite{icml24FairGrad}. Hybrid methods combine both families~\cite{liu2021IMTL, linreasonableTMLR2022, liuautoTMLR2022},  while others operate on gradients w.r.t.\ shared features to reduce cost or semantic disparity~\cite{sener2018multi, Javaloy2021RotoGradGH, wang2025repmtl}.

\vspace{-0.60em}
\paragraph{Multi-Task Model Architectures.}
(i)~\textit{CNN-based models}: 
promote cross-task sharing through multimodal distillation~\cite{xu2018pad}, affinity-based task pairing~\cite{zhang2019PAP}, multi-scale interaction~\cite{vandenhende2020MTINet}, and learnable fusion across task-specific streams~\cite{misra2016CrossStitch, ruder2019SluiceNet, gao2019nddr}. However, CNNs are constrained to local receptive fields, which motivates the shift to global attention.
(ii)~\textit{Transformer-based models}:
Transformers~\cite{vaswani2017attention} offer global modeling and scalable pretrained features, both benefiting MTL. InvPT~\cite{ye2022invpt, ye2024invplpl} employs cross-attention for joint modeling of global positions and multiple tasks.
MTFormer~\cite{xu2022mtformer} applies cross-task attention with contrastive learning, MQTransformer~\cite{xu2023MQTransformer} proposes query attention to model inter-task dependencies. DeMT~\cite{xu2023demt} combines deformable convolutions with query-based Transformers via shared gating. While effective, these architectures focus on increasingly complex decoders without designs for task-intrinsic priors~\cite{zhong2024transforming}.
(iii)~\textit{MoE and state-space models}:
Mixture-of-experts (MoE) naturally suit MTL by routing shared experts in a task-conditioned manner~\cite{ye2023taskexpert, chen2023mod, yang2024MLoRE}, though they mainly target decoder components and lack global cross-task sharing. Recent state-space models such as MTMamba~\cite{lin2024mtmamba} and PAME~\cite{yu2026pame} leverage selective scanning for linear-complexity cross-task interaction, but remain rely on specially-tailored encoder or decoder designs for multi-task learning.

Across all these lines, enriching task interaction typically requires modifying the encoder's internals or introducing complex decoders, coupling methods to specific backbones and hindering adaptation of off-the-shelf VFMs~\cite{Oquab2023DINOv2, simeoni2025dinov3}. VTP departs from this by operating alongside the vision encoder, preserving full VFM compatibility while providing sufficient capacity for expressive task prompting.

\vspace{-0.60em}
\paragraph{Adapters and Prompting.}
MTLoRA~\cite{agiza2024mtlora} disentangles shared and task-specific parameters via low-rank adapters. Visual Prompt Tuning~\cite{jia2022VPT, yoo2023improving} introduces learnable prompts into ViT layers for task transfer, with extensions to language and graphs~\cite{he2022hyperprompt, yu2024multigprompt}. In the multi-task setting, a growing line of work employs parameter-efficient fine-tuning (PEFT) by inserting lightweight modules into frozen model backbones for dense prediction~\cite{bhattacharjee2023vitAdapterMTL}, including hypernetwork-generated adapters~\cite{liu2022polyhistor}, shared projections with task-specific extraction~\cite{xin2024vmt}, task-conditional adapters~\cite{liang2023visualExemplar, jiang2024task}, prompt adaptation or tuning~\cite{shen2024MTVLPT}, input-conditioned filters~\cite{baek2025tadformer}, and diffeomorphic singular-value modulation~\cite{mantri2025ditask}. Among these studies, multi-task prompting-based architectures~\cite{ye2023taskprompter, lu2024PGT, herzig2024promptonomyvit} are most related to VTP, as they adopt task prompting with ViTs for joint training. We discuss their limitations that inspire VTP in Sec.~\ref{subsec:prelim}. Unlike the above PEFT and adaptation methods that freeze the backbone and prioritize parameter efficiency, VTP as a multi-task framework jointly trains the entire model with task vectors and targets multi-task performance through adaptive read-write routing of a richer virtual task state.

\section{Methodology}
\label{sec:method}

This section presents Virtual Task Prompting (VTP). We first introduce the overall framework of multi-task prompting architectures, covering the encoding and decoding pipelines with notation in Sec.~\ref{subsec:prelim}. We then discuss VTP's task-adaptive read-and-write routing mechanism, which maintains a $T{+}V$ virtual state alongside the encoder and feeds into a streamlined decoder in Sec.~\ref{subsec:vtp}. Finally, we conclude with the multi-task joint training objective after VTP's output prediction in Sec.~\ref{subsec:joint_training}.

\subsection{Overall Framework}
\label{subsec:prelim}

\paragraph{Encoding.}
Given an input RGB image, a patch embedding layer produces $N{=}H{\times}W$ visual tokens  $\mathcal{X}^{0}=\{x^{0}_1, x^{0}_2, \dots, x^{0}_N\}\in\mathbb{R}^{N\times C}$, where $H$ and $W$ are spatial resolution after patchifying and $C$ denotes the embedding dimension. To jointly address $T$ correlated tasks, a set of $T$ learnable vectors $\mathcal{P}^{0}=\{p^{0}_1, p^{0}_2, \dots, p^{0}_T\}\in\mathbb{R}^{T\times C}$ is introduced, with one for each task. The task vectors and visual tokens are concatenated to derive a joint input sequence $\mathcal{Z}^{0}$, as illustrated in Figure~\ref{fig:VTP}:
\begin{equation}
    \mathcal{Z}^{0} = \left[\mathcal{P}^{0};\; \mathcal{X}^{0}\right] = \left[p^{0}_1, \dots, p^{0}_T;\; x^{0}_1, \dots, x^{0}_N\right] \;\in\; \mathbb{R}^{(T+N)\times C},
    \label{eq:token_sequence}
\end{equation}
where $[\,\cdot\;;\;\cdot\,]$ denotes token-wise concatenation. The joint sequence is processed through $L$ ViT~\cite{dosovitskiy2020image} layers. At each model layer $l\in\{1,\dots,L\}$, an encoder block $\mathcal{F}_l (\cdot)$, typically comprising multi-head self-attention, a feed-forward network, and residual connections, updates the joint sequence as:
\begin{equation}
    \left[\mathcal{P}^{l};\;\mathcal{X}^{l}\right] = \mathcal{F}_l\!\left(\left[\mathcal{P}^{l-1};\;\mathcal{X}^{l-1}\right]\right),
    \label{eq:vit_block}
\end{equation}
where $\mathcal{P}^{l}\in\mathbb{R}^{T\times C}$ and $\mathcal{X}^{l}\in\mathbb{R}^{N\times C}$ indicate the output task vectors and image tokens within layer $l$, respectively. Self-attention over all $T{+}N$ tokens yields a $(T{+}N)\times(T{+}N)$ affinity map, allowing each task to absorb relevant cues from the input image and to exchange information with other tasks.

\vspace{-0.60em}
\paragraph{Decoding.}
After the final layer $L$, the processed task vectors $\mathcal{P}^{L}=\{p^{L}_t\}_{t=1}^{T}$ and visual tokens $\mathcal{X}^L$ drive task-specific predictions. For each task $t\in\{1,\dots,T\}$, we define the task-specific affinities $\mathcal{M}^{L} \in \mathbb{R}^{T\times N}$ (colored orange in Figure~\ref{fig:VTP}(c)), where each row $\mathcal{M}^{L}_{t,:}$ is computed as: 
\begin{equation}
    \mathcal{M}^{L}_{t,:} = \mathrm{softmax}\!\left(\frac{\mathcal{X}^{L}\,{p^{L}_t}^{\!\top}}{\sqrt{C}}\right) \;\in\; \mathbb{R}^{N},
    \label{eq:affinity}
\end{equation}
which captures how task vector $p^{L}_t$ attends to the image and highlights the regions that are most relevant to task $t$. Similarly, we can derive the cross-task affinities $\mathcal{R}^{L} \in \mathbb{R}^{T\times T}$ (blue in Figure~\ref{fig:VTP}(c)), encoding inter-task interactions during training. The decoder $\mathcal{D}_t (\cdot, \cdot, \cdot)$ then produces output $\hat{\mathcal{Z}}_t$ as:
\begin{equation}
    \hat{\mathcal{Z}}_t = \mathcal{D}_t\!\left(\mathcal{M}_{t,:}^{L},\; \mathcal{R}_{t,:}^{L},\; \mathcal{X}^{L}\right)\;\in\; \mathbb{R}^{N \times C},
    \label{eq:decoding}
\end{equation}
where $\mathcal{M}^{L}_{t,:}$ routes shared features to task $t$ and $\mathcal{R}^{L}_{t,:}$ retrieves complementary information from other tasks based on their affinities. These two operations encode task-specific and agnostic information in a prompt-guided manner, followed by a lightweight convolutional head for the final predictions.

\vspace{-0.60em}
\paragraph{Limitations.}
This framework combines the parameter efficiency of HPS with the task sharing of SPS. However, (i) each task is represented by a single fixed vector updated solely via backpropagation, which limits its expressiveness (with analysis in Sec~\ref{subsec:exp_analysis}). (ii) To compensate, prior work~\cite{ye2023taskprompter} splits task vectors to spatial and channel ones in Eq.~\eqref{eq:token_sequence} for disentangled processing, inducing modifications to both the encoder $\mathcal{F}_l(\cdot)$ in Eq.~\eqref{eq:vit_block} (\textit{e.g.}, extra channel attention) and decoder $\mathcal{D}_t(\cdot, \cdot, \cdot)$ in Eq.~\eqref{eq:decoding} (\textit{e.g.}, dual-branch prompting). As shown in Figure~\ref{fig:VTP}(a), such re-engineering couples the method to specific encoder structures, inevitably hindering its compatibility with off-the-shelf pretrained VFMs.

\subsection{Virtual Task Prompting}
\label{subsec:vtp}

The limitations above point to key demands for multi-task prompting architectures. (i)~The encoder $\mathcal{F}_l(\cdot)$ should stay clean to preserve compatibility with VFMs. (ii)~A single vector per task provides insufficient representation capacity. A natural remedy is to increase its scale, yet as shown in Table~\ref{tab:ablation}, doubling or quadrupling vectors yields negligible or even negative gains. Our analysis in Sec.~\ref{subsec:exp_analysis} further shows that the vectors undergo severe mode collapse across layers, converging to near-identical non-discriminative patterns. This motivates a third, (iii)~an end-to-end trainable approach to encode and optimize task vectors rather than treating them as static weights updated via backpropagation.

We address these with Virtual Task Prompting (VTP), which operates outside model backbone as a bypass before and after each ViT block. Concretely, VTP introduces $V$ additional vectors as \textit{virtual tasks} alongside the encoder, enlarging the embedding space to $T{+}V$. To achieve discriminative task representations, we propose a read–and-write mechanism for \textit{virtual tasks} inspired by hyper-connections~\cite{zhu2024hyperconnections, li2025virtual, xie2025mhc}, where learnable routing matrices adaptively read $T$ compositions from the $T{+}V$ virtual state at each layer $l$ for interaction with image tokens as in Eq.~\ref{eq:vit_block}, then write the results back, providing explicit learning pathways and more expressive task features across different layers.

\subsubsection{Encoding}
\label{subsubsec:vtp_encoding}

As shown in Figure~\ref{fig:VTP}(b), virtual task state $\mathcal{P}^{0}\in\mathbb{R}^{(T+V)\times C}$ are initialized as learnable embeddings which operate outside the encoder. Their roles are not prescribed but optimized in an end-to-end manner. Some may encode shared structure across correlated tasks, while others capture task-specific patterns. At each ViT layer $l\in\{1,\dots,L\}$, VTP executes a read–and–write cycle as follows.

\textbf{State Read.}\; To ensure the encoder receive $T$ task vectors in the standard $(T{+}N)\times C$ dimension, a learnable routing matrix $\Phi^{l}\in\mathbb{R}^{T\times(T+V)}$ accesses virtual state $\mathcal{P}^{l-1}$ by reading $T$ vectors from it:
\begin{equation}
    \tilde{\mathcal{P}}^{l} = \Phi^{l}\,\mathcal{P}^{l-1} \;\in\; \mathbb{R}^{T\times C},
    \label{eq:read}
\end{equation}
where each read vector $\tilde{p}^{l}_t$ is a learned task-specific composition over the virtual state $\mathcal{P}^{l}$, encoding richer information than a static vector per task. Note that different tasks may draw on shared states for common structure or access distinct ones for specialization, which is determined by the learnable $\Phi^{l}$. The $T$ vectors are then concatenated with image tokens and processed by encoder $\mathcal{F}_l (\cdot)$ as:
\begin{equation}
    \left[\hat{\mathcal{P}}^{l};\;\mathcal{X}^{l}\right] = \mathcal{F}_l\!\left(\left[\tilde{\mathcal{P}}^{l};\;\mathcal{X}^{l-1}\right]\right),
    \label{eq:process}
\end{equation}
where $\hat{\mathcal{P}}^{l}\in\mathbb{R}^{T\times C}$ and $\mathcal{X}^{l}\in\mathbb{R}^{N\times C}$ are the processed task vectors and visual tokens. Because the Read step compresses the virtual state to exactly $T$ vectors, the input retains the standard shape $\mathbb{R}^{(T+N)\times C}$, and $\mathcal{F}_l$ — self-attention, feed-forward network, and residual connections — remains entirely unchanged, preserving direct compatibility with off-the-shelf VFM weights.

\textbf{State Write.}\; After the encoder processes $T$ vectors, two learnable routing matrices write the updated information back into $T{+}V$ virtual state to maintain the expanded capacity across layers as:
\begin{equation}
    \mathcal{P}^{l} = \Psi^{l}\,\hat{\mathcal{P}}^{l} \;+\; \Gamma^{l}\,\mathcal{P}^{l-1} \;\in\; \mathbb{R}^{(T+V)\times C},
    \label{eq:write}
\end{equation}
where $\Psi^{l}\in\mathbb{R}^{(T+V)\times T}$ writes the encoder-processed task vectors into selected virtual slots, and $\Gamma^{l}\in\mathbb{R}^{(T+V)\times(T+V)}$ carries forward information from the previous virtual state $\mathcal{P}^{l-1}$. The two terms balance what to update from the current layer with what to retain from preceding ones, enabling virtual task vectors to accumulate and populate discriminative task information across layers.

As such, VTP imposes task-adaptive selection to the $T{+}V$ vectors that distills which task patterns are essential enough for encoder interaction, counteracting the mode collapse that arises when vectors are updated through backpropagation alone. Throughout this cycle, the routing operates only on the task-vector stream with $\mathcal{O}((T+V)^2 C)$ complexity independent of $N$. Since $T, V \ll N$ in dense scene understanding, this routing overhead is negligible relative to the encoder cost (as in Sec.~\ref{subsec:exp_efficiency}).

\subsubsection{Decoding}
\label{subsubsec:vtp_decoding}

Since VTP operates on unsplit task vectors, the decoding is simplified compared to the dual-branch design of TaskPrompter~\cite{ye2023taskprompter}. As illustrated in Figure~\ref{fig:VTP}(c), for each task $t$, we got its task-specific affinity $\mathcal{M}^{L}_{t,:}$ (Eq.~\ref{eq:affinity}) and cross-task affinity $\mathcal{R}^{L}_{t,:}$ during training, then decode the output $\hat{\mathcal{Z}}_t$ as:
\begin{equation}
    \hat{\mathcal{Z}}_t = \mathcal{D}_t\!\left(\mathcal{M}_{t,:}^{L},\; \mathcal{R}_{t,:}^{L},\; \mathcal{X}^{L}\right)
    = \sum_{k=1}^{T} \mathcal{R}_{t,k}^{L} \;\big(\underbrace{\mathcal{M}_{k,:}^{L} \odot \mathcal{X}^{L}}_{\text{task-specific}}\big),
    \label{eq:vtp_decoding}
\end{equation}
where the inner term extracts task-specific features for each task $k$ by prompting with $\mathcal{M}_{k,:}^{L}$, and the outer summation aggregates these task-specific features across tasks with the cross-task affinities $\mathcal{R}_{t,:}^{L}$, allowing each task to retrieve complementary information from other tasks. Note that each row of the output is a weighted combination of all tasks' feature maps. A lightweight prediction head, a $3{\times}3$ convolutional block followed by a linear projection, is then applied to $\hat{\mathcal{Z}}_t$ to produce the final prediction for task $t$. Following prior work~\cite{ye2022invpt, ye2023taskprompter}, we select the output features at $K$ evenly spaced encoder layers and apply the same prompting-based decoding process in Eq.~\ref{eq:vtp_decoding} to each of them.

\begin{figure}[h]
    \centering
    \includegraphics[width=\textwidth]{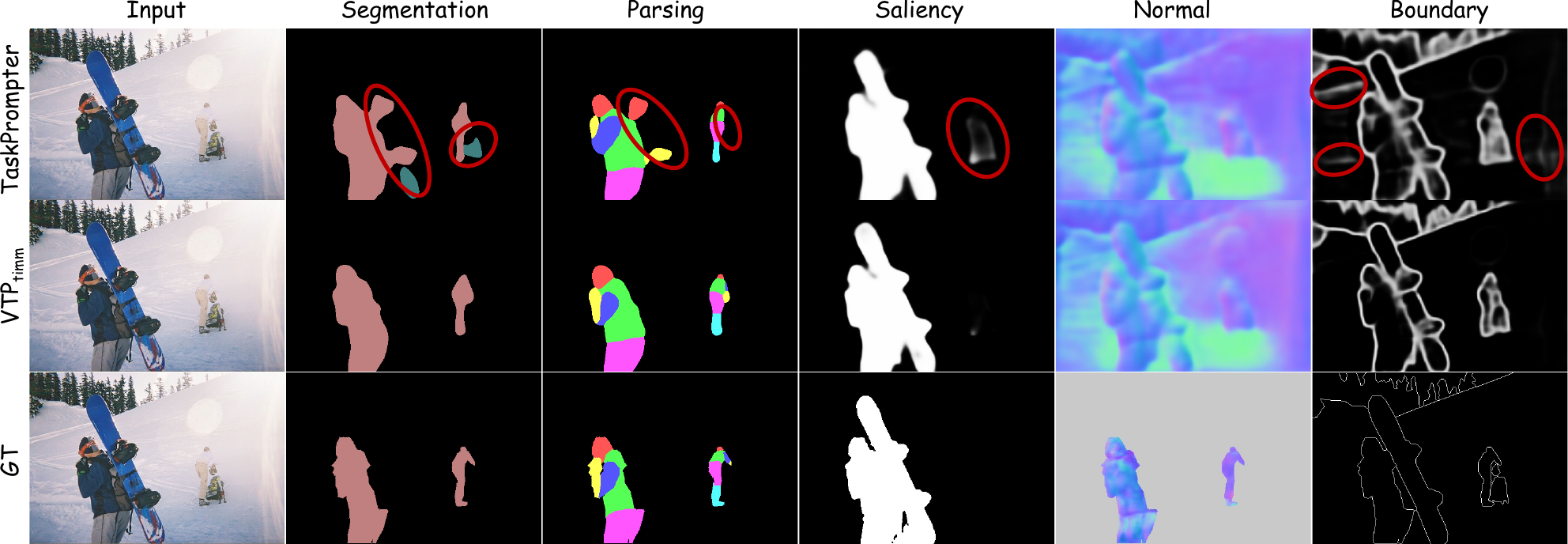}
     \vspace{-1.50em}
        \caption{\textbf{Qualitative results on Pascal-Context~\cite{chen2014PascalContext}.} VTP\textsubscript{timm} produces precise fine-grained object segmentation (bags), body-part parsing (human arms) with tighter saliency (suppress the spurious background). Key regions are with red circles. VTP\textsubscript{vit-5} and VTP\textsubscript{dinov3} and more results on Figure~\ref{fig:pascal_vis_full}.}
\vspace{-0.40em}
    \label{fig:pascal_vis_short}
\end{figure}

\begin{figure}[h]
    \centering
    \includegraphics[width=\textwidth]{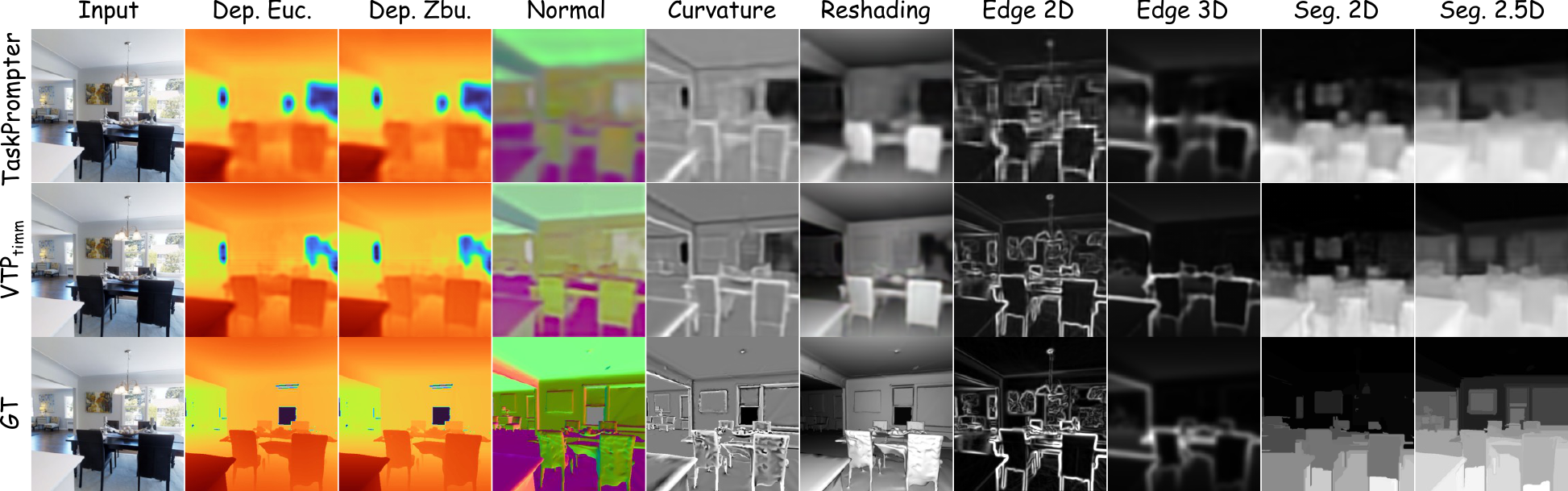}
     \vspace{-1.50em}
        \caption{\textbf{Qualitative results on Taskonomy~\cite{zamir2018taskonomy} with 9 dense tasks.} VTP\textsubscript{timm} produces apparently more accurate predictions than the baseline. VTP\textsubscript{vit-5} and VTP\textsubscript{dinov3} and more results on Figure~\ref{fig:taskonomy_qual}. }
    \label{fig:tasko_vis_short}
\end{figure}

\subsection{Joint Training}
\label{subsec:joint_training}
VTP comprises three main components: (i)~a standard ViT encoder $\mathcal{F}_l$ that processes both image and task tokens, (ii)~the read-write matrices that maintains and routes virtual states as a bypass, and (iii)~a streamlined decoder $\mathcal{D}_t$ with prediction heads.
We combine per-task objectives into a single loss:
\begin{equation}
    \mathcal{L}_\text{MTL}=\sum_{t=1}^{T} \lambda_{t} \cdot \mathcal{L}_{t}(\theta_{s},\, \theta_{t}),
    \label{eq:total_loss}
\end{equation}
where $\theta_{s}$ denotes shared encoder parameters, $\theta_{t}$ the decoder parameters for task $t$, and $\lambda_{t}$ scalar weights balancing each task's contribution. Since all task losses backpropagate through the shared encoder, VTP routing matrices $\Phi^{l}$, $\Psi^{l}$, and $\Gamma^{l}$ receive gradient signals from every task simultaneously, shaping the read–write pathways to reflect cross-task structure in an end-to-end learnable manner.

\section{Experiments}
\label{sec:exp}
We evaluate VTP on three MTL benchmarks, covering both indoor and outdoor scenarios: (i) Pascal-Context~\cite{chen2014PascalContext} (5 tasks, outdoor scenes), (ii) NYUD-v2~\cite{silberman2012NYUv2} (4 tasks, indoor scenes), and (iii) Taskonomy~\cite{zamir2018taskonomy} (13 heterogeneous tasks, stress test). We then present ablations (Sec.~\ref{subsec:exp_ablation}), qualitative results (Figure~\ref{fig:pascal_vis_short}), and empirical analysis of learned task vectors  (Sec.~\ref{subsec:exp_analysis}) and efficiency (Sec.~\ref{subsec:exp_efficiency}).

\begin{table*}[!t]
\centering
\caption{\textbf{State-of-the-art comparison} on NYUD-v2~\cite{silberman2012NYUv2} (left) and PASCAL-Context~\cite{chen2014PascalContext} (right). All Transformer-based models use a ViT-L backbone. VTP$_\text{timm}$ is with ImageNet-pretrained weights, while VTP$_\text{dinov3}$ and VTP$_\text{vit-5}$ employ DINOv3~\cite{simeoni2025dinov3} and ViT-5~\cite{wang2026vit5} without any modification. $\Delta_p$ (\%) denotes the average MTL gains relative to TaskPrompter~\cite{ye2023taskprompter} ($\Delta_p{=}0$). $\uparrow (\downarrow)$ indicates higher (lower) values are better. The best and second-best results per task are in \textbf{bold} and \underline{underline}, respectively.}
\label{tab:sota}

\setlength{\tabcolsep}{4pt}
\renewcommand{\arraystretch}{1.05}

\resizebox{\linewidth}{!}{%
\begin{tabular}{@{}l cccc c@{}}
\toprule
\multirow{2}{*}{\textbf{Method}} & \textbf{Semseg} & \textbf{Depth} & \textbf{Normal} & \textbf{Edge} & {$\Delta_p$} \\
& mIoU\,$\uparrow$ & RMSE\,$\downarrow$ & mErr\,$\downarrow$ & odsF\,$\uparrow$ & (\%)\,$\uparrow$ \\
\midrule
\multicolumn{6}{@{}l}{\cellcolor{gray94}\textit{ \textbf{CNN-based}}} \\
Cross-Stitch~\cite{misra2016CrossStitch} & 36.34 & 0.6290 & 20.88 & 76.38 & $-$17.94 \\
PAP~\cite{zhang2019PAP} & 36.72 & 0.6178 & 20.82 & 76.42 & $-$17.13 \\
PSD~\cite{zhou2020PSD} & 36.69 & 0.6246 & 20.87 & 76.42 & $-$17.54 \\
PAD-Net~\cite{xu2018pad} & 36.61 & 0.6270 & 20.85 & 76.38 & $-$17.68 \\
MTI-Net~\cite{vandenhende2020MTINet} & 45.97 & 0.5365 & 20.27 & 77.86 & $-$7.80 \\
ATRC~\cite{bruggemann2021ATRC} & 46.33 & 0.5363 & 20.18 & 77.94 & $-$7.48 \\
\midrule
\multicolumn{6}{@{}l}{\cellcolor{gray94}\textit{ \textbf{Transformer-based}}} \\
InvPT~\cite{ye2022invpt} & 53.56 & 0.5183 & 19.04 & 78.10 & $-$1.74 \\
InvPT++~\cite{ye2024invplpl} & 53.85 & 0.5096 & 18.67 & 78.10 & $-$0.69 \\
TaskPrompter~\cite{ye2023taskprompter} & 55.30 & 0.5152 & 18.47 & 78.20 & 0.00 \\
TaskExpert~\cite{ye2023taskexpert} & 55.35 & 0.5157 & 18.54 & 78.40 & $-$0.03 \\
MQTransformer~\cite{xu2023MQTransformer} & 54.84 & 0.5325 & 19.67 & 78.20 & $-$2.67 \\
TSP-Transformer~\cite{wang2024tsp} & 55.39 & 0.4961 & 18.44 & 77.50 & +0.78 \\
MLoRE~\cite{yang2024MLoRE} & 55.96 & 0.5076 & \underline{18.33} & 78.43 & +0.93 \\
\midrule
\multicolumn{6}{@{}l}{\cellcolor{gray94}\textit{ \textbf{Other architectures}}} \\
MTMamba~\cite{lin2024mtmamba} & 55.82 & 0.5066 & 18.63 & \underline{78.70} & +0.60 \\
TaskDiffusion~\cite{yang2025multi} & 55.65 & 0.5020 & 18.43 & 78.64 & +0.99 \\
PAME~\cite{yu2026pame} & 56.63 & 0.5102 & 18.52 & 78.10 & +0.74 \\
\midrule
\rowcolor{vtpblue!90} VTP$_\text{timm}$ & \underline{57.21} & 0.4960 & 18.41 & 78.43 & +1.95 \\
\rowcolor{vtpblue!90} VTP$_\text{vit-5}$ & 56.58 & \underline{0.4915} & \textbf{18.18} & 78.65 & \underline{+2.27} \\
\rowcolor{vtprose!90} VTP$_\text{dinov3}$ & \textbf{59.55} & \textbf{0.4820} & 18.52 & \textbf{79.25} & \textbf{+3.80}   \\
\bottomrule
\end{tabular}%
\hspace{2em}%
\begin{tabular}{@{}l ccccc c@{}}
\toprule
\multirow{2}{*}{\textbf{Method}} & \textbf{Semseg} & \textbf{Parsing} & \textbf{Saliency} & \textbf{Normal} & \textbf{Edge} & {$\Delta_p$} \\
& mIoU\,$\uparrow$ & mIoU\,$\uparrow$ & maxF\,$\uparrow$ & mErr\,$\downarrow$ & odsF\,$\uparrow$ & (\%)\,$\uparrow$ \\
\midrule
\multicolumn{7}{@{}l}{\cellcolor{gray94}\textit{ \textbf{CNN-based}}} \\
ASTMT~\cite{maninis2019attentive} & 68.00 & 61.10 & 65.70 & 14.70 & 72.40 & $-$11.69 \\
PAD-Net~\cite{xu2018pad} & 53.60 & 59.60 & 65.80 & 15.30 & 72.50 & $-$16.51 \\
MTI-Net~\cite{vandenhende2020MTINet} & 61.70 & 60.18 & 84.78 & 14.23 & 70.80 & $-$8.76 \\
ATRC~\cite{bruggemann2021ATRC} & 62.69 & 59.42 & 84.70 & 14.20 & 70.96 & $-$8.67 \\
ATRC-ASPP~\cite{bruggemann2021ATRC} & 63.60 & 60.23 & 83.91 & 14.30 & 70.86 & $-$8.57 \\
ATRC-BMTAS~\cite{bruggemann2021ATRC} & 67.67 & 62.93 & 82.29 & 14.24 & 72.42 & $-$6.65 \\
\midrule
\multicolumn{7}{@{}l}{\cellcolor{gray94}\textit{ \textbf{Transformer-based}}} \\
InvPT~\cite{ye2022invpt} & 79.03 & 67.61 & 84.81 & 14.15 & 73.00 & $-$1.60 \\
InvPT++~\cite{ye2024invplpl} & 80.22 & 69.12 & 84.74 & 13.73 & 74.20 & +0.06 \\
TaskPrompter~\cite{ye2023taskprompter} & 80.89 & 68.89 & 84.83 & 13.72 & 73.50 & 0.00 \\
TaskExpert~\cite{ye2023taskexpert} & 80.64 & 69.42 & 84.87 & 13.56 & 73.30 & +0.28 \\
MQTransformer~\cite{xu2023MQTransformer} & 78.93 & 67.41 & 83.58 & 14.21 & 73.90 & $-$1.81 \\
TSP-Transformer~\cite{wang2024tsp} & 81.48 & 70.64 & 84.86 & 13.69 & 74.80 & +1.06 \\
MLoRE~\cite{yang2024MLoRE} & 81.41 & 70.52 & 84.90 & 13.51 & 75.42 & +1.45 \\
\midrule
\multicolumn{7}{@{}l}{\cellcolor{gray94}\textit{ \textbf{Other architectures}}} \\
MTMamba~\cite{lin2024mtmamba} & 81.11 & 72.62 & 84.14 & 14.14 & \textbf{78.80} & +1.80 \\
TaskDiffusion~\cite{yang2025multi} & 81.21 & 69.62 & 84.94 & 13.55 & 74.89 & +0.94 \\
PAME~\cite{yu2026pame} & 81.38 & 70.39 & 84.51 & \underline{13.36} & 75.30 & +1.50 \\
\midrule
\rowcolor{vtpblue!90} VTP$_\text{timm}$ & \underline{82.88} & \underline{73.85} & 85.52 & 13.56 & 75.12 & +2.77 \\
\rowcolor{vtpblue!90} VTP$_\text{vit-5}$ & 82.48 & 73.28 & \underline{85.60} & \textbf{13.32} & 75.35 & \underline{+2.94} \\
\rowcolor{vtprose!90} VTP$_\text{dinov3}$ & \textbf{85.62} & \textbf{79.25} & \textbf{85.96} & 13.85 & \underline{76.80} & \textbf{+5.15} \\
\bottomrule
\end{tabular}%
}

\end{table*}

\subsection{Datasets and Evaluation}
\label{subsec:dataset}

\paragraph{Datasets.}
(i)~\textbf{Pascal-Context}~\cite{chen2014PascalContext} contains 4,998/5,105 train/test images labeled for semantic segmentation (Semseg), human parsing (Parsing), and boundary detection (Edge).  Following previous studies~\cite{maninis2019attentive}, we use pseudo labels for surface normal estimation (Normal) and saliency detection (Saliency), totaling 5 tasks.
(ii)~\textbf{NYUD-v2}~\cite{silberman2012NYUv2} provides 1,449/795 train/test images with 4 tasks spanning semantics and geometry: Semseg, monocular depth estimation (Depth), Normal, and Edge.
(iii)~\textbf{Taskonomy}~\cite{zamir2018taskonomy} comprises 381.8K images with 25/5 buildings for train/test for its tiny split, with 13 tasks: object and scene classification (Obj./Scn.\ Cls.), Euclidean and z-buffer depth (Dep.\ Euc./Zbu.), normal, curvature (Curv.), reshading (Reshad.), 2D/3D edge and keypoint (Edge/Keyp.\ 2D/3D), and semseg (Seg.\ 2D/2.5D). It is challenging for jointly training heterogeneous tasks.


\begin{table*}[!t]
\centering
\caption{\textbf{Taskonomy results over 13 tasks.} $\Delta_p$ (\%) denotes the average MTL performance gains relative to TaskPrompter~\cite{ye2023taskprompter} baseline ($\Delta_p{=}0$). $\uparrow (\downarrow)$ indicates the higher (lower) values are better. The best and second-best results for each task are highlighted in \textbf{bold} and \underline{underline}, respectively.}
\label{tab:taskonomy}

\setlength{\tabcolsep}{3.5pt}
\renewcommand{\arraystretch}{1.05}

\resizebox{\textwidth}{!}{%
\begin{tabular}{@{}l cc cc ccc cccc cc c@{}}
\toprule
\multirow{2}{*}{\textbf{Method}}
& \textbf{Obj.Cls.} & \textbf{Scn.Cls.} & \textbf{Dep.Euc.} & \textbf{Dep.Zbu.} & \textbf{Normal} & \textbf{Curv.} & \textbf{Reshad.} & \textbf{Edge\,2D} & \textbf{Edge\,3D} & \textbf{Keyp.\,2D} & \textbf{Keyp.\,3D} & \textbf{Seg.\,2D} & \textbf{Seg.\,2.5D}
& {$\Delta_p$} \\
& Acc\,$\uparrow$ & Acc\,$\uparrow$ & RMSE\,$\downarrow$ & RMSE\,$\downarrow$ & mErr\,$\downarrow$ & L2\,$\downarrow$ & L1\,$\downarrow$ & L1\,$\downarrow$ & L1\,$\downarrow$ & L1\,$\downarrow$ & L1\,$\downarrow$ & L1\,$\downarrow$ & L1\,$\downarrow$
& (\%)\,$\uparrow$ \\
\midrule
InvPT~\cite{ye2022invpt} & 57.01 & 65.50 & 6.251 & 6.384 & 29.850 & 0.9120 & 0.1810 & 0.0985 & \underline{0.0835} & 0.1150 & 0.0835 & 0.2035 & 0.1590 & $-$2.34 \\
TaskPrompter~\cite{ye2023taskprompter} & 58.37 & 66.76 & 6.150 & 6.268 & 29.046 & 0.8947 & 0.1754 & 0.0961 & \textbf{0.0819} & 0.1107 & 0.0814 & \underline{0.1991} & \underline{0.1566} & 0.00 \\
\midrule
\rowcolor{vtpblue!90} VTP$_\text{timm}$ & 58.78 & \textbf{70.96} & 5.757 & 5.823 & \textbf{27.505} & 0.8650 & 0.1750 & 0.0804 & 0.0948 & 0.0834 & 0.0805 & \textbf{0.1927} & 0.1570 & +4.50 \\
\rowcolor{vtpblue!90} VTP$_\text{vit-5}$ & \textbf{61.73} & \underline{69.90} & \underline{5.460} & \textbf{5.620} & 33.060 & \underline{0.8494} & \underline{0.1703} & \underline{0.0795} & 0.0862 & \textbf{0.0710} & \underline{0.0796} & 0.2020 & 0.1608 & \underline{+5.53} \\
\rowcolor{vtprose!90} VTP$_\text{dinov3}$ & \underline{60.21} & 68.08 & \textbf{5.440} & \underline{5.630} & \underline{28.356} & \textbf{0.8288} & \textbf{0.1447} & \textbf{0.0683} & 0.0859 & \underline{0.0719} & \textbf{0.0747} & 0.1992 & \textbf{0.1515} & \textbf{+9.58} \\  
\bottomrule
\end{tabular}}

\end{table*}

\vspace{-0.60em}
\paragraph{Evaluation.}
Following established protocols~\cite{ye2023taskprompter, ye2022invpt}, we report standard per-task metrics: mIoU (Semseg, Parsing), mErr (Normal), maxF/odsF (Edge), Acc (classification), and $L_1$/$L_2$ distances (regression). Depth is tested with RMSE, absolute/relative error, and threshold ($1.25$, $1.25^2$, $1.25^3$). To summarize overall MTL performance, we report the popular average per-task relative gain $\Delta_p$:

\begin{align}
\Delta_p =\frac{1}{T} \sum_{t=1}^{T}(-1)^{\sigma_{t}} \frac{M_{m, t}-M_{b, t}} {M_{b, t}} 
\label{eq:delta_p}
\end{align}
where $M_{m,t}$ and $M_{b,t}$ are performance of method $m$ and the baseline on task $t$, and $\sigma_t \in \{0, 1\}$ flips the sign for lower-is-better metrics (\textit{e.g.}, mErr, $L_1$, $L_2$). Please refer to
Appendix for more details.

\vspace{-0.60em}
\paragraph{Implementation Details.}
Following prior work~\cite{ye2022invpt, ye2023taskprompter}, all models are trained for $40{,}000$ iterations using . Adam with learning rate $2 \times 10^{-5}$, weight decay $1 \times 10^{-6}$, a batch size of $4$ and a polynomial schedule. Regression tasks are trained with $L_1$ loss, while classification tasks (Semseg, Parsing, Saliency, Edge) using cross-entropy. Task vectors are initialized from $\mathcal{N}(\mu{=}1, \sigma{=}1)$. $\Phi^l$ and $\Gamma^l$ are initialized as block identity matrices to ensure pass-through reading and full carry-forward, while $\Psi^l$ is with a cyclic pattern ($\Psi^l_{i,j}{=}1$ if $i{=}j\!\bmod T$, else $0$) so that each task writes back to its natural state, reducing to standard prompting for stability, with routing structure emerging during training.

\subsection{Pascal-Context and NYUD-v2 Results}
\label{subsec:exp_scene_bench}
Table~\ref{tab:sota} compares VTP against CNN-based, Transformer-based, and other architectures on Pascal-Context~\cite{chen2014PascalContext} and NYUD-v2~\cite{silberman2012NYUv2}. We report three VTP variants: VTP$_\text{timm}$ with ImageNet-pretrained ViT, and VTP$_\text{dinov3}$ / VTP$_\text{vit-5}$ that integrate DINOv3~\cite{simeoni2025dinov3} and ViT-5~\cite{wang2026vit5} without any modification.

\vspace{-0.60em}
\paragraph{VTP without VFMs.}
VTP$_\text{timm}$ surpasses existing methods on $\Delta_p$ on both benchmarks. On Pascal-Context, it achieves $+2.77\%$ (previous best: $+1.80\%$), leading on segmentation ($82.88$ vs.\ $81.48$) and parsing ($73.85$ vs.\ $72.62$), while also improving saliency ($85.52$ maxF). On NYUD-v2, it reaches $+1.95\%$, nearly doubling the prior best of $+0.99\%$, with the highest segmentation ($57.21$ vs.\ $56.63$) results. Boundary detection on Pascal-Context remains below MTMamba ($78.80$), likely due to the Mamba-based decoder that excels at fine-grained spatial details. Overall, these results confirm that VTP's adaptive prompting is effective on its own, independent of stronger pretrained vision encoders.

\vspace{-0.60em}
\paragraph{VTP with VFMs.}
VFM integration yields a substantial leap, showcasing the practical value of VTP's clean encoder. On Pascal-Context, VTP$_\text{dinov3}$ achieves $+5.15\%$, nearly doubling VTP$_\text{timm}$, with segmentation rising from $82.88$ to $85.62$ and parsing from $73.85$ to $79.25$.On NYUD-v2, it reaches $+3.80\%$, roughly $4\times$ the prior non-VTP best, leading on segmentation ($59.55$), depth ($0.4820$), and boundary ($79.25$). VFMs show diverse strengths: VTP$_\text{vit-5}$ achieves the best normal results ($18.18$, $13.32$), while VTP$_\text{dinov3}$ dominates semantic tasks. Interestingly, VTP$_\text{dinov3}$ slightly degrades normal estimation compared to VTP$_\text{timm}$ ($13.85$ vs.\ $13.56$ on Pascal-Context; $18.52$ vs.\ $18.41$ on NYUD-v2), suggesting that DINOv3's self-supervised pretraining favors semantic over geometric representations.

\subsection{Stress-Testing on Taskonomy}
\label{subsec:exp_tiny_taskonomy}
To examine VTP's robustness under extreme MTL conditions, we evaluate on Taskonomy~\cite{zamir2018taskonomy} with 13 heterogeneous tasks, from sparse classification (object and scene) to dense regression (curvature, reshading). Due to limited compute and the absence of popular benchmarking, we train and compare against two baselines~\cite{ye2022invpt, ye2023taskprompter}. The goal here is not to establish a new leaderboard, but to stress-test whether our VTP with different vision encoders scales gracefully as task count and diversity grow.

As shown in Table~\ref{tab:taskonomy}, VTP$_\text{timm}$ achieves $+4.50\%$ over TaskPrompter, with notable gains appearing on scene classification ($70.96$ vs.\ $66.76$), depth ($5.757$ vs.\ $6.150$), normals ($27.505$ vs.\ $29.046$), and 2D edges ($0.0804$ vs.\ $0.0961$). VTP$_\text{dinov3}$ reaches $+9.58\%$ overall, more than doubling VTP$_\text{timm}$, with the best results on 7 of 13 tasks including depth, curvature, and reshading. VTP$_\text{vit-5}$ achieves $+5.53\%$ with a complementary profile, leading on object classification and 2D keypoints. Not all tasks benefit uniformly. VTP$_\text{timm}$ regresses on 3D edges ($0.0948$ vs.\ $0.0819$), and VTP$_\text{vit-5}$ degrades on normals ($33.060$ vs.\ $29.046$) despite leading this on Pascal-Context and NYUD-v2, likely due to distributional differences in Taskonomy's annotations. Despite these variations, the consistent overall gains across 13 tasks of varying granularity show that VTP generalizes well beyond the common 4--5 task setting.

\begin{figure}[t]
    \centering
    \includegraphics[width=\textwidth]{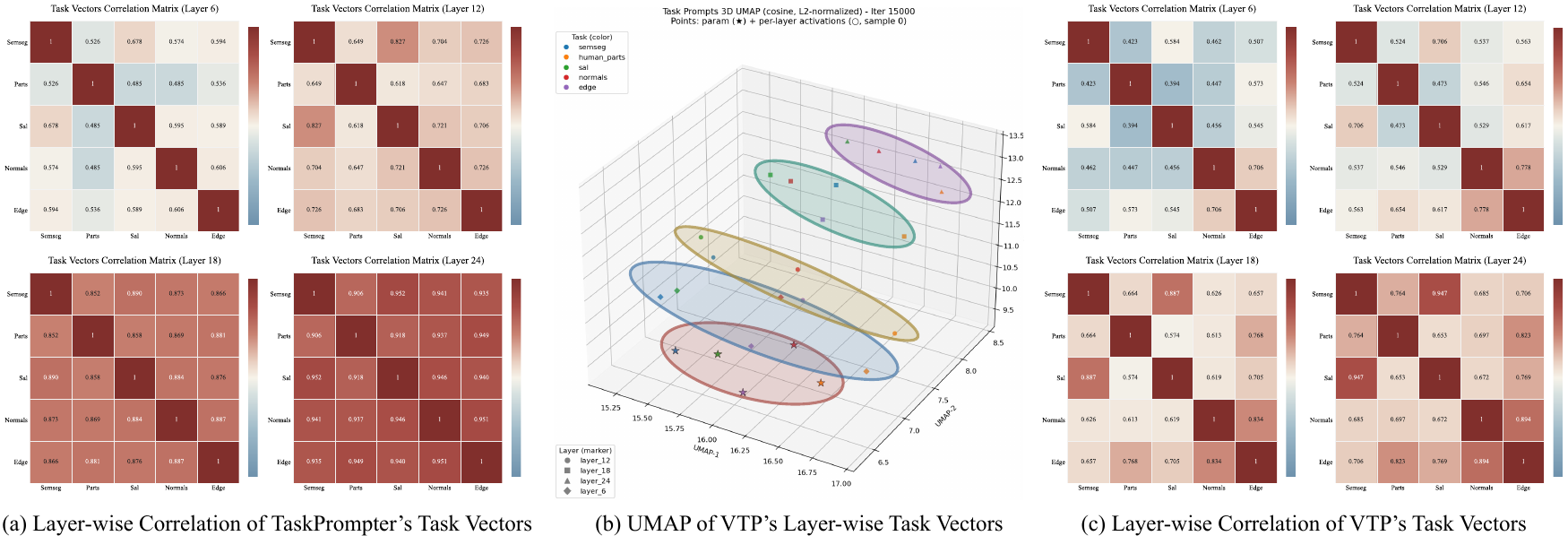}
    \vspace{-0.80em}
    \caption{\textbf{Task correlation and UMAP analysis.}
Pairwise cosine similarity among task vectors at selected layers for \textbf{(a)} TaskPrompter collapses into uniformly 
high correlation with negligible variation in deep layers. \textbf{(b)}~UMAP of VTP's task vectors shows clear layer-wise clusters, confirming that vectors form an internally cohesive latent space encoding layer-wise discriminative information. \textbf{(c)}~VTP 
exhibits structured task affinities, capturing adaptive inter-task 
differentiation across depths.}
\vspace{-1.20em}
    \label{fig:umap}
\end{figure}

\subsection{Qualitative results and analysis}
\label{subsec:exp_analysis}
Figure~\ref{fig:pascal_vis_short} compares VTP with TaskPrompter on Pascal-Context (more in Appendix~\ref{sec:app_qualitative}), where VTP$_\text{timm}$ yields sharper boundaries and VFM variants further refine details in Figure~\ref{fig:pascal_vis_full}. To understand how these task vectors evolve across layers, we analyze pairwise cosine similarity of task vectors and visualize them via UMAP~\cite{mcinnes2018umap} in Figure~\ref{fig:umap}. Please refer to Appendix~\ref{sec:app_umap} for more setting details.

\vspace{-0.80em}
\paragraph{Mode collapse of the baseline.} In Figure~\ref{fig:umap}~(a), TaskPrompter's task vectors progressively over-align with depth. While shallow layers exhibit moderate, structured inter-task similarity, this erodes rapidly. The mean off-diagonal correlation reaches $0.87$ with a min--max range of $0.04$ by layer~18 and $0.94$ by layer~24. This indicates that, without explicit learning pathways beyond pure backpropagation, task vectors degenerate into near-identical directions regardless of their specific semantic relationships.

\vspace{-0.80em}
\paragraph{Discriminative task patterns of VTP.} In contrast, VTP exhibits structured, depth-related patterns (Figure~\ref{fig:umap}(c)). Shallow layers correlate geometric tasks (normals, edges), while deeper layers increasingly align semantic pairs (segmentation and saliency), showing that VTP encodes discriminative features from the $T{+}V$ virtual state into $T$ vectors across depths. Empirical analysis in Figure~\ref{fig:umap}~(b) further corroborates this observation, where different layers establish their distinct UMAP clusters.

\subsection{Ablations and Hyperparameter Sensitivity}
\label{subsec:exp_ablation}

\paragraph{Ablation Study.}
Table~\ref{tab:ablation}(a) isolates the gains from added capacity and VTP itself. Doubling decoder parameters slightly hurts $\Delta_p$ ($-0.18\%$), and naively increasing vectors yields negligible or negative returns ($+0.09\%$ for $2\times$, $-0.14\%$ for $4\times$). In contrast, VTP ($V{=}2$) achieves $+2.77\%$, confirming that gains stem from the method more than increased capacity.
Table~\ref{tab:ablation}(b) varies the virtual task $V$. Even $V{=}0$ offers $+1.52\%$, since it includes the full Eqs.~\ref{eq:read}--\ref{eq:write}, showing that layer-wise routing alone is crucial, consistent with the mode collapse in Figure~\ref{fig:umap}. Adding virtual tasks contributes: $V{=}2$ yields $+0.97\%$ over $V{=}0$, while $V{=}4$ and $V{=}8$ perform comparably ($+2.41\%$, $+2.43\%$), showing that a small $V$ suffices for capacity. We use $V{=}2$ as default. We conduct more validations in Table~\ref{tab:appendix_comparison}.


\begin{table}[!t]
\centering
\caption{\textbf{Ablation (left) and efficiency (right) results.} Left: $\Delta_{pascal}$ and $\Delta_{nyu}$ denote the MTL gains $\Delta_{p}$ (\%) on Pascal-Context and NYUD-v2 relative to TaskPrompter~\cite{ye2023taskprompter}. $V$: virtual task number. VTPs share similar parameters and FLOPs. Throughput results are measured on a single L40 GPU.}
\label{tab:ablation}
\vspace{-0.06in}

\setlength{\tabcolsep}{4pt}
\renewcommand{\arraystretch}{1.08}

\resizebox{0.94\columnwidth}{!}{%
\begin{tabular}{@{}l c c@{}}
\toprule
\textbf{Configuration} & $\Delta_{pascal}$ & $\Delta_{nyu}$ \\
\midrule
\multicolumn{3}{@{}l}{\textit{\color{gray}(a) Capacity controls}} \\
TaskPrompter & 0.00 & 0.00 \\
\quad + 2$\times$ decoder parameters & $-$0.18 & $-$0.07 \\
\quad + 2$\times$ vectors per task & +0.09 & +0.05 \\
\quad + 4$\times$ vectors per task & $-$0.14 & $-$0.02 \\
\midrule
\multicolumn{3}{@{}l}{\textit{\color{gray}(b) Virtual task number $V$}} \\
VTP ($V{=}0$) & +1.52 & +1.28 \\
\grow VTP ($V{=}2$, default) & \textbf{+2.77} & \underline{+1.95} \\
VTP ($V{=}4$) & +2.41 & \textbf{+1.97} \\
VTP ($V{=}8$) & \underline{+2.43} & +1.84 \\
\bottomrule
\end{tabular}%
\hspace{1.5em}%
\begin{tabular}{@{}l cccc@{}}
\toprule
\textbf{Method} & $\Delta_p$ & \textbf{Parameters} & \textbf{FLOPs} & \textbf{Throughput} \\
& (\%)\,$\uparrow$ & (M)\,$\downarrow$ & (G)\,$\downarrow$ & (imgs/s)\,$\uparrow$ \\
\midrule
PAD-Net~\cite{xu2018pad}          & $-$16.51 & \textbf{330} & 773 & 7.684 \\
MTI-Net~\cite{vandenhende2020MTINet} & $-$8.76 & 851 & 774 & 6.790 \\
InvPT~\cite{ye2022invpt}          & $-$1.60 & 423 & 669 & 10.813 \\
MQTransformer~\cite{xu2023MQTransformer} & $-$1.81 & 504 & 712 & 9.045 \\
TaskExpert~\cite{ye2023taskexpert} & +0.28 & 420 & 622 & 11.526 \\
TaskPrompter~\cite{ye2023taskprompter} & 0.00 & 401 & 497 & 14.394 \\
\midrule
\rowcolor{vtpblue!90} VTP$_\text{timm}$ & +2.77 & 342 & \underline{487} & \textbf{16.318} \\
\rowcolor{vtpblue!90} VTP$_\text{vit-5}$ & \underline{+2.94} & \underline{337} & \textbf{487} & 13.258 \\
\rowcolor{vtprose!90} VTP$_\text{dinov3}$ & \textbf{+5.15} & 336 & 488 & \underline{16.153} \\
\bottomrule
\end{tabular}}
\vspace{-0.40em}
\end{table}

\vspace{-0.60em}
\paragraph{Efficiency.}
\label{subsec:exp_efficiency}
The complexity comprises three parts: (i) Self-attention over $T+N$ tokens ($\mathcal{O}((T+N)^2 C)$, scaling with $N$); (ii) VTP routing on task stream ($\mathcal{O}((T+V)^2 C)$, independent of $N$); and (iii) decoder affinities ($\mathcal{O}(T(N+T)C)$, linear in $N$). Since $T, V \ll N$, the routing cost is negligible relative to encoder when $T, V \ll N$. 
Table~\ref{tab:ablation} (right) shows that VTP uses fewer parameters (342M vs.\ 401M) and FLOPs (487G vs.\ 497G) than TaskPrompter with higher throughput, as it removes extra encoder designs. Thus, VTP variants share similar costs, with differences stemming from the VFMs.

\section{Conclusion and Discussion}
\label{sec:conclusion}

\vspace{-0.50em}
\paragraph{Conclusion.}
We introduce Virtual Task Prompting (VTP), a prompting-based multi-task model that eliminates the need for modifications to the vision encoder. By preserving a clean backbone, VTP integrates well with off-the-shelf VFMs, further pushing the limits on Pascal-Context and NYUD-v2. Stress testing on Taskonomy with 13 diverse tasks also shows VTP's robustness and scalability.

\vspace{-0.80em}
\paragraph{Limitations and Future Work.}
Due to limited resources, we have not tried VTP at larger multimodal foundation models like Qwen3-VL~\cite{bai2025qwen3vl}. Exploring expressive prompting methods and scaling up modalities would be beneficial for applications like autonomous driving and embodied AI.

\begin{ack}
The work is supported in part by the Early Career Scheme of the Research Grants Council (RGC) of the Hong Kong SAR under grant No. 26202321, ITF PRP/046/24FX, SAIL Research Project, and HKUST-Zeekr Collaborative Research Fund.
\end{ack}

\bibliography{ref}
\bibliographystyle{abbrv}


\clearpage
\appendix

\section{Appendix}
\label{sec:app_outline}
This appendix provides additional implementation details, quantitative results, qualitative visualizations, and further discussions that complement the main text. It is organized as follows:

In Appendix~\ref{sec:app_dataset}, we provide the dataset and implementation specifications for Pascal-Context~\cite{chen2014PascalContext}, NYUD-v2~\cite{silberman2012NYUv2}, and Taskonomy~\cite{zamir2018taskonomy}, covering task definitions, evaluation metrics, data augmentation, optimization schedules, loss functions, task balancing weights, and hyperparameter settings.

In Appendix~\ref{sec:app_taskonomy_quant}, we present the complete results on Taskonomy~\cite{zamir2018taskonomy}, reporting the full suite of depth estimation metrics (RMSE, absolute error, relative error, and threshold accuracies $\delta_1$, $\delta_2$, $\delta_3$) for both Euclidean and z-buffer depth, which were omitted from the main text due to the space limitations.

In Appendix~\ref{sec:app_qualitative}, we provide extended qualitative comparisons beyond Figures~\ref{fig:pascal_vis_short}, highlighting challenging scenarios such as thin structures, articulated poses, and crowded multi-person layouts, as well as one set of Taskonomy~\cite{zamir2018taskonomy} predictions spanning nine typical dense prediction tasks.

In Appendix~\ref{sec:app_umap}, we present extended activation correlation analysis and UMAP~\cite{mcinnes2018umap} visualizations of the learned task embeddings. We show that the prompts in TaskPrompter exhibit a mode collapse at deep layers, while VTP's prompts form clearly separated, task-discriminative clusters in UMAP subspace and demonstrate progressive specialization across different network layers.

In Appendix~\ref{sec:app_flexibility}, we formalize VTP's flexibility and drop-in compatibility with off-the-shelf VFMs, and compare these properties against previous model architectures for multi-task scene understanding.

\section{Dataset and Implementation Details}
\label{sec:app_dataset}
 
In this section, we discuss the task definitions, evaluation metrics, and training protocols for our three evaluation benchmarks, Pascal-Context, NYUD-v2, and Taskonomy. These implementation details complement the overview in Sec.~\ref{subsec:dataset} of the main text and are intended to facilitate reproducibility.
 
\subsection{Pascal-Context}
\label{sec:app_pascal}
 
Pascal-Context~\cite{chen2014PascalContext} extends the PASCAL VOC 2010 dataset with dense per-pixel annotations for the entire scene, covering both indoor and outdoor imagery. Following the standard protocol from previous studies~\cite{ye2022invpt, ye2023taskprompter, ye2024invplpl}, we use 4,998 images for training and 5,105 images for testing. The five tasks jointly trained on this benchmark are:
\textbf{Semantic Segmentation (Semseg):} 21-class per-pixel labeling (20 object categories plus background). Evaluated with mean Intersection over Union (mIoU, \%). \textbf{Human Parsing (Parsing):} 7-class dense segmentation of human body parts, including head, torso, upper arms, lower arms, upper legs, and lower legs. Evaluated with mIoU (\%). \textbf{Saliency Detection (Saliency):} Binary per-pixel saliency estimation using the pseudo-labels generated by~\cite{maninis2019attentive}. Evaluated with the maximal F-measure (maxF). \textbf{Surface Normal Estimation (Normal):} Per-pixel prediction of 3D surface orientation vectors. Following~\cite{maninis2019attentive}, we use the pseudo-labels produced by a pretrained normal estimator. Evaluated with mean angular error (mErr, in degrees; lower is better). \textbf{Boundary Detection (Edge or Boundary):} Binary edge detection using the DistillBDCN~\cite{maninis2019attentive} annotations. Evaluated with the optimal-dataset-scale F-measure (odsF).
 
All images are resized to $512 \times 512$ during training and testing, consistent with prior work~\cite{ye2022invpt, ye2023taskprompter}. Data augmentation includes random horizontal flipping, random scaling (scale factor sampled uniformly from $[0.5, 2.0]$), and random cropping. No test-time augmentation is applied. For the training configuration on Pascal-Context, we use the Adam optimizer with an initial learning rate of $2 \times 10^{-5}$ and weight decay of $1 \times 10^{-6}$. A polynomial learning rate schedule with power $0.9$ is applied over 40,000 iterations with a batch size of 4. The task-balancing weights $\lambda_t$ in Eq.~\ref{eq:total_loss} are set the same as previous studies~\cite{ye2023taskprompter, wang2025repmtl}.
For VTP-specific modules, we set the default number of virtual tasks $V = 2$ (see Table~\ref{tab:ablation} in the main text for sensitivity analysis). All learnable task vector embeddings, including the virtual ones, are randomly initialized from a normal distribution with $\mathcal{N}(\mu{=}1, \sigma{=}1)$. $\Phi^l$ and $\Gamma^l$. The VTP read-write routing matrices $\Phi^l$ and $\Gamma^l$ are initialized as block identity matrices to ensure pass-through reading and full carry-forward, while $\Psi^l$ is with a cyclic pattern ($\Psi^l_{i,j}{=}1$ if $i{=}j\!\bmod T$, else $0$) so that each task writes back to its natural state, reducing to standard prompting for stability, with routing structure emerging during training.

\subsection{NYUD-v2}
\label{sec:app_nyud}
 
NYUD-v2~\cite{silberman2012NYUv2} is a widely used indoor scene understanding benchmark consisting of RGB-D images captured with a Microsoft Kinect sensor. Following the standard protocol~\cite{ye2022invpt, ye2023taskprompter}, we use 1,449 images for training and 795 images for testing. The four tasks are: \textbf{Semantic Segmentation (Semseg):} 40-class per-pixel labeling of indoor scene categories. Evaluated with mIoU (\%). \textbf{Monocular Depth Estimation (Depth):} Per-pixel prediction of absolute depth from a single RGB input. Evaluated with root mean squared error (RMSE; lower is better). \textbf{Surface Normal Estimation (Normal):} Per-pixel prediction of 3D surface normals. Evaluated with mean angular error (mErr, in degrees; lower is better). \textbf{Boundary Detection (Edge or Boundary):} Object boundary detection. Evaluated with the optimal-dataset-scale F-measure (odsF).
 
Images are resized to $512 \times 512$ during both training and testing. The data augmentation strategy is identical to Pascal-Context: random horizontal flipping, random scaling with a scale factor in $[0.5, 2.0]$, and random cropping. No test-time augmentation is used.
The training setup on NYUD-v2 mirrors Pascal-Context in most respects: Adam optimizer with learning rate $2 \times 10^{-5}$, weight decay $1 \times 10^{-6}$, polynomial schedule with power $0.9$, 40,000 iterations, and batch size 4. The task-balancing weights $\lambda_t$ in Eq.~\ref{eq:total_loss} are set the same as previous studies~\cite{ye2023taskprompter, wang2025repmtl}. All VTP hyperparameters ($V = 2$, VTP read-write matrices initialization, and multi-scale extraction at $K = 4$ blocks) remain identical to Pascal-Context. The prompts are randomly initialized with normal distribution. All experiments are implemented with PyTorch and run on NVIDIA L40-48G GPUs.

\subsection{Taskonomy}
\label{sec:app_taskonomy}
 
Taskonomy~\cite{zamir2018taskonomy} is a large-scale indoor dataset originally comprising approximately 4.5 million images from over 500 buildings, annotated for 26 diverse tasks. To make experimentation tractable under our computational constraints, we adopt the official \textit{tiny} split, which contains 381.8K images sampled from 35 buildings (25 for training and 5 for testing, with the remaining 5 reserved for validation). From the full set of 26 tasks, we select 13 representative ones that span a broad spectrum of visual understanding: \textbf{Classification tasks (sparse supervision):} \textit{Object Classification (Obj.~Cls.):} Image-level prediction of the dominant object category. Evaluated with top-1 accuracy (Acc, \%). \textit{Scene Classification (Scn.~Cls.):} Image-level prediction of the scene type (\textit{e.g.}, bedroom, kitchen, office). Evaluated with top-1 accuracy (Acc, \%). \textbf{Regression tasks (dense supervision):} \textit{Euclidean Depth (Dep.~Euc.):} Per-pixel Euclidean distance from the camera. Evaluated with RMSE (lower is better). \textit{Z-buffer Depth (Dep.~Zbu.):} Per-pixel depth along the camera's optical axis. Evaluated with RMSE (lower is better). \textit{Surface Normals (Normal):} Per-pixel 3D surface orientation. Evaluated with mean angular error (mErr; lower is better). \textit{Curvature (Curv.):} Per-pixel principal curvature estimation. Evaluated with L2 distance (lower is better). \textit{Reshading (Reshad.):} Per-pixel re-rendering under novel illumination. Evaluated with L1 distance (lower is better). \textit{2D Edge Detection (Edge 2D):} Binary edge maps from RGB cues. Evaluated with L1 distance (lower is better). \textit{3D Edge Detection (Edge 3D):} Occluding contour detection from geometric cues. Evaluated with L1 distance (lower is better). \textit{2D Keypoint Detection (Keyp.~2D):} Per-pixel keypoint heatmap estimation. Evaluated with L1 distance (lower is better). \textit{3D Keypoint Detection (Keyp.~3D):} Geometric keypoint detection in 3D. Evaluated with L1 distance (lower is better). \textit{2D Segmentation (Seg.~2D):} Unsupervised 2D region segmentation. Evaluated with L1 distance (lower is better). \textit{2.5D Segmentation (Seg.~2.5D):} Unsupervised segmentation incorporating depth cues. Evaluated with L1 distance (lower is better).
 
This selection is deliberately diverse: it includes both image-level classification and pixel-level dense prediction, spans semantic, geometric, and low-level structural understanding, and encompasses tasks with markedly different output distributions and gradient magnitudes. This diversity makes Taskonomy an ideal stress test for multi-task architectures, as the risk of negative transfer grows substantially with the number and heterogeneity of tasks.
 
All images are resized to $224 \times 224$ during training and testing, following the resolution convention of the original Taskonomy benchmark and previous studies~\cite{zamir2018taskonomy, chen2023mod}. Data augmentation consists of random horizontal flipping and random color jittering (brightness, contrast, saturation, and hue), without random scaling or cropping due to the already moderate resolution.
Our training uses the Adam optimizer with an initial learning rate of $1 \times 10^{-5}$ (lower than the other benchmarks to accommodate the larger dataset and greater task diversity), weight decay of $1 \times 10^{-6}$, and a polynomial learning rate schedule with power $0.9$ over 80,000 iterations (doubled relative to Pascal-Context and NYUD-v2 to allow sufficient convergence across 13 tasks) with a batch size of 8 due to the limited computational resource. Classification tasks use cross-entropy loss, while all dense regression tasks use L1 loss. The task-balancing weights $\lambda_t$ are set uniformly to 1.0 for all 13 tasks for impartial comparisons. All VTP hyperparameters and VTP read-write matrices initialization remain unchanged, with multi-scale features extracted at $K = 4$ evenly spaced blocks. All experiments are implemented with PyTorch and run on NVIDIA L40-48G GPUs.

\section{Full Quantitative Results on Taskonomy}
\label{sec:app_taskonomy_quant}
 
As discussed in Section~\ref{subsec:dataset}, depth estimation is evaluated with a richer set of metrics beyond RMSE, including absolute error (Abs), relative error (Rel), and threshold accuracy ($\delta_1 = 1.25$, $\delta_2 = 1.25^2$, $\delta_3 = 1.25^3$). While Table~2 of the main text reports only RMSE for conciseness, we present the full depth evaluation for both Euclidean depth and z-buffer depth on Taskonomy in Table~\ref{tab:taskonomy_depth}.
 
\begin{table}[h]
\centering
\caption{\textbf{Full depth estimation metrics on Taskonomy.} We report RMSE, absolute error (Abs), relative error (Rel), and threshold accuracies ($\delta_1$, $\delta_2$, $\delta_3$) for both Euclidean depth and z-buffer depth. $\downarrow$ indicates lower is better; $\uparrow$ indicates higher is better. Best and second-best results are marked in \textbf{bold} and \underline{underline}, respectively.}
\label{tab:taskonomy_depth}
\vspace{-0.05in}
\resizebox{\textwidth}{!}{%
\begin{tabular}{l|cccccc|cccccc}
\toprule
& \multicolumn{6}{c|}{\textbf{Euclidean Depth (Dep.~Euc.)}} & \multicolumn{6}{c}{\textbf{Z-buffer Depth (Dep.~Zbu.)}} \\
\cmidrule(lr){2-7} \cmidrule(lr){8-13}
Method & RMSE$\downarrow$ & Abs$\downarrow$ & Rel$\downarrow$ & $\delta_1 \uparrow$ & $\delta_2 \uparrow$ & $\delta_3 \uparrow$
       & RMSE$\downarrow$ & Abs$\downarrow$ & Rel$\downarrow$ & $\delta_1 \uparrow$ & $\delta_2 \uparrow$ & $\delta_3 \uparrow$ \\
\midrule
InvPT~\cite{ye2022invpt}
  & 6.251 & 4.873 & 0.412 & 34.52 & 60.18 & 75.24
  & 6.384 & 4.965 & 0.420 & 33.87 & 59.43 & 74.61 \\
TaskPrompter~\cite{ye2023taskprompter}
  & 6.150 & 4.784 & 0.398 & 36.10 & 61.85 & 76.43
  & 6.268 & 4.871 & 0.406 & 35.42 & 61.06 & 75.78 \\
\midrule
\rowcolor{vtpblue!90} VTP\textsubscript{timm}
  & 5.757 & 4.402 & \underline{0.353} & 39.85 & 65.23 & 79.16
  & 5.823 & 4.456 & 0.374 & 39.17 & 64.52 & 78.48 \\
\rowcolor{vtpblue!90} VTP\textsubscript{vit-5}
  & \underline{5.460} & \underline{4.118} & {0.358} & \underline{42.73} & \underline{67.95} & \underline{81.30}
  & \underline{5.620} & \underline{4.262} & \underline{0.352} & \underline{41.55} & \underline{66.88} & \underline{80.42} \\
\rowcolor{vtprose!90} VTP\textsubscript{dinov3}
  & \textbf{5.440} & \textbf{4.093} & \textbf{0.337} & \textbf{43.15} & \textbf{68.40} & \textbf{81.72}
  & \textbf{5.630} & \textbf{4.240} & \textbf{0.348} & \textbf{41.92} & \textbf{67.25} & \textbf{80.85} \\
\bottomrule
\end{tabular}%
}
\end{table}
 
Several observations are worth noting. First, VTP consistently outperforms both InvPT and TaskPrompter across all depth metrics on both depth variants. The improvements are not confined to a single error measure: VTP reduces RMSE, absolute error, and relative error simultaneously, while substantially boosting all three threshold accuracies. This indicates that the gains are not artifacts of a particular evaluation criterion but reflect genuinely improved depth predictions across the full error distribution.
 
Second, VFM variants (VTP\textsubscript{vit-5} and VTP\textsubscript{dinov3}) offer large improvements on the threshold accuracy metrics. For instance, on Euclidean depth, VTP\textsubscript{dinov3} improves $\delta_1$ from 36.10\% (TaskPrompter) to 43.15\%, a gain of 7.05 absolute points. This suggests that VFM features are especially beneficial for reducing large outlier errors, as the threshold metrics are sensitive to the fraction of pixels with predictions within a fixed multiplicative factor of the ground truth.
Third, the two VFMs exhibit a similar pattern to what we observed in the main text: VTP\textsubscript{dinov3} achieves the best overall depth performance on most metrics, while VTP\textsubscript{vit-5} is competitive and occasionally matches or exceeds DINOv3 on specific measures (\textit{e.g.}, z-buffer Rel). This supports the finding that different VFMs bring complementary strengths, and VTP can leverage each of them as a plug-and-play backbone.

\section{Additional Qualitative Comparisons}
\label{sec:app_qualitative}
 
\begin{figure}[t]
    \centering
    \includegraphics[width=\textwidth]{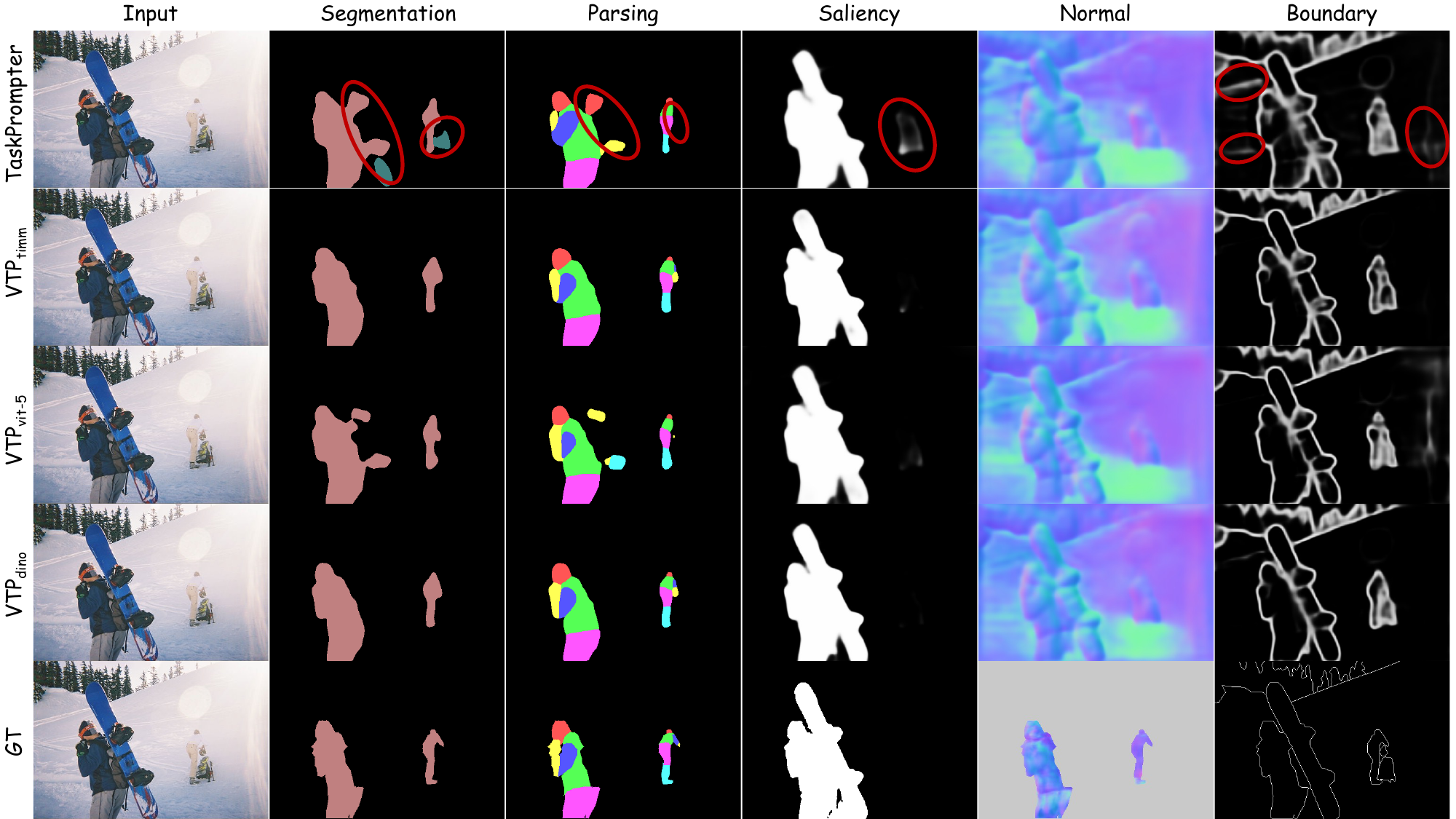}
     \vspace{-1.50em}
        \caption{\textbf{Qualitative results on Pascal-Context~\cite{chen2014PascalContext}.} VTP\textsubscript{timm} produces precise fine-grained object segmentation (bags), body-part parsing (human arms) with tighter saliency (suppress the spurious background). VTP\textsubscript{vit-5} and VTP\textsubscript{dinov3} 
further sharpen the details. Key regions are with red circles.}
\vspace{-0.40em}
    \label{fig:pascal_vis_full}
\end{figure}

To complement the qualitative comparison in Figure~\ref{fig:pascal_vis_short} of the main manuscript, we present extra qualitative results that further illustrate VTP's advantages over the previous multi-task prompting method TaskPrompter~\cite{ye2023taskprompter} across different scenes, tasks, and difficulty levels. We first provide the full version of Figure~\ref{fig:pascal_vis_short} for apple-to-apple comparison in Figure~\ref{fig:pascal_vis_full}. Set~1 in Figure~\ref{fig:pascal_qual_2} focuses on multi-person scenes where inter-individual occlusion and proximity create severe ambiguity for parsing and segmentation. Set~2 in Figure~\ref{fig:pascal_qual_2} targets a scene with extremely small and thin semantic objects alongside dense, complicated edge structures. The Taskonomy visualization in Figure~\ref{fig:taskonomy_qual} demonstrates VTP's effectiveness in the extreme MTL setting, emphasizing the progressive quality improvement across VTP variants requiring no additional method integration efforts.

\subsection{Additional Pascal-Context Results (Set 1)}
 
\begin{figure}[t]
    \centering
    \includegraphics[width=\textwidth]{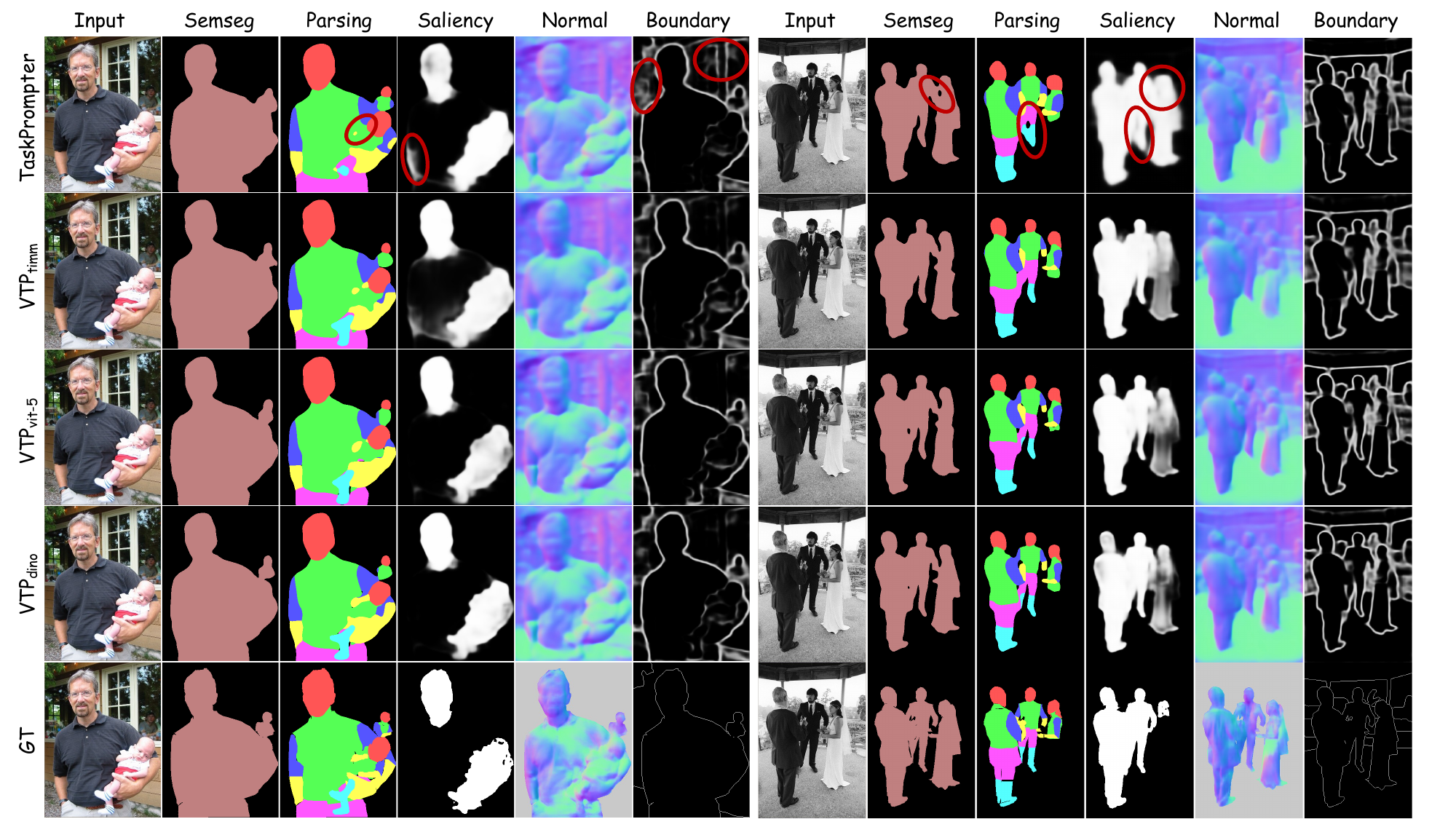}
    \caption{\textbf{Additional qualitative results on Pascal-Context (Set 1).} Two multi-person outdoor scenes that stress-test inter-individual disambiguation.  \textbf{Left:} a man holding an infant against a cluttered natural background. 
TaskPrompter~\cite{ye2023taskprompter} (red circles) lets semantic labels bleed 
across the person with background contour in segmentation, 
fragments body-part assignments at the overlapping arms in parsing, 
and over-highlights background clutter in boundary. \textbf{Right:} a crowded group where closely spaced individuals share similar 
appearance. TaskPrompter's parsing labels entangle across adjacent persons, 
with leg assignments bleeding from one individual into the next. In both scenes, VTP\textsubscript{timm} recovers coherent per-person 
segmentation masks, disentangled body-part labels, and tighter saliency focus. 
VTP\textsubscript{vit-5} and VTP\textsubscript{dinov3}~\cite{wang2026vit5, simeoni2025dinov3} 
further resolve fine-grained inter-person boundaries, saliency, and produce 
geometrically smoother surface normals, approaching the ground-truth.}
    \label{fig:pascal_qual_2}
\end{figure}

Figure~\ref{fig:pascal_qual_2} shows predictions on two additional Pascal-Context scenes, both featuring multiple human subjects—a scenario that is particularly demanding for multi-task scene understanding models because it requires simultaneously resolving per-person semantic segmentation, fine-grained body-part parsing, instance-aware saliency, geometrically consistent surface normals, and precise boundary delineation across overlapping and mutually occluding individuals.
 
The left scene depicts a close-up outdoor portrait with a man holding an infant against a complex natural background. As highlighted by the red circles, TaskPrompter produces segmentation errors at the boundary between the person and the background, where semantic categories bleed across the contour. Its parsing output also exhibits fragmented body-part assignments with ambiguous region boundaries. In contrast, VTP\textsubscript{timm} already delivers noticeably cleaner segmentation and saliency masks with tighter adherence to the true object silhouette. TaskPrompter over-highlights background clutter near the person, while VTP concentrates its saliency response more precisely on the foreground subject. Moving to VTP\textsubscript{vit-5} and VTP\textsubscript{dinov3}, the predictions progressively sharpen, with DINOv3~\cite{simeoni2025dinov3}'s strong semantic representations yielding parsing maps that faithfully delineate individual body parts with minimal inter-part confusion.
 
The right scene also presents a hard case: a group of multiple people walking together in a crowded outdoor setting, where inter-person occlusion and similar appearance create substantial ambiguity. TaskPrompter struggles most visibly here in the parsing task—the body-part labels for adjacent and partially overlapping individuals become entangled, with limb assignments frequently bleeding across person boundaries (see red circles). The segmentation output also shows confusion at the interfaces between closely spaced people. As shown in Figure~\ref{fig:pascal_qual_2}, VTP handles this multi-person scenario markedly better: even VTP\textsubscript{timm} produces more coherent per-person region assignments, and the VFM-enhanced variants further resolve the fine-grained inter-person boundaries. The normal estimation is another notable differentiator: VTP generates spatially smooth and geometrically plausible normal fields over each person's body surface, whereas TaskPrompter's normals exhibit noisy, blurred, locally inconsistent orientations at precisely the regions where multiple people overlap.
 
These two multi-person scenes confirm that VTP provides the representational capacity needed to jointly resolve instance-level ambiguities across tasks, where the single-prompt-per-task baseline is fundamentally capacity-limited.

\subsection{Additional Pascal-Context Results (Set 2)}
 
Figure~\ref{fig:pascal_qual_2} presents a particularly challenging scene on Pascal-Context~\cite{chen2014PascalContext} that stress-tests multi-task models along two orthogonal axes simultaneously: (i)~extremely small and thin semantic objects that occupy only a handful of pixels, and (ii)~dense, complicated edge structures that arise from the rich spatial layout of the scene. This combination is especially demanding as the model must preserve fine-grained spatial detail for accurate segmentation and parsing of tiny objects while also producing dense, high-fidelity boundary maps across the entire image.

\begin{figure}[t]
    \centering
    \includegraphics[width=\textwidth]{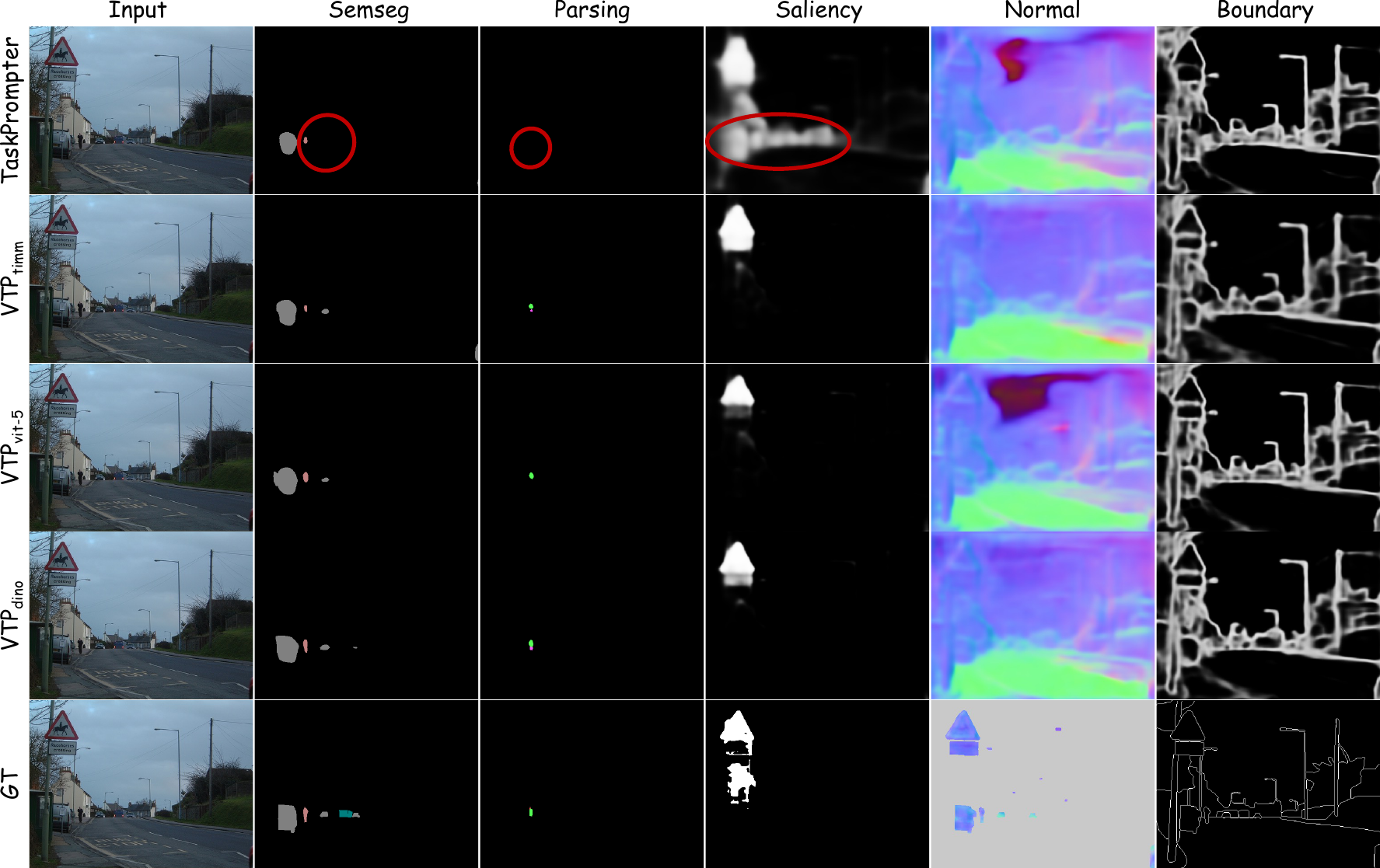}
    \caption{\textbf{Additional qualitative results (Set 2).} A challenging street scene that stress-tests two orthogonal challenges simultaneously: 
extremely small and thin semantic objects (distant poles, a tiny person occupying 
only a handful of pixels) and dense edge structures from road markings, 
building contours, and lampposts. TaskPrompter~\cite{ye2023taskprompter} (red circles) drops these tiny objects 
entirely from its segmentation and parsing maps, and produces a large 
spurious saliency blob on the background sky. VTP\textsubscript{timm} begins to recover the thin structures in 
segmentation and parsing while suppressing the false saliency response. VTP\textsubscript{vit-5} and VTP\textsubscript{dinov3}
further sharpen the objects (the blue and green masks in Semseg) and produce boundaries that 
capture both macro building contours and micro pole edges.}
    \label{fig:pascal_qual_3}
\end{figure}

The scene depicts a road with distant structures, lampposts, poles, a small human object, and other slender objects set against a backdrop of buildings and sky. These thin structures pose a severe challenge for segmentation: a single misclassified pixel can cause an entire pole or post to vanish from the prediction, or merge with the surrounding background class. As highlighted by the red circles, TaskPrompter drops these small objects altogether—tiny human objects are either entirely absent from its segmentation and parsing maps. VTP\textsubscript{timm}, even without VFM features, already begins to recover these thin objects. This improvement suggests that the richer prompt representations afforded by the virtual tasks provide the discriminative power necessary to preserve spatial detail that a single static prompt vector simply cannot encode. The boundary detection task reveals an equally striking contrast. The ground truth for this scene contains a dense web of edges arising from road markings, building contours, poles, and background textures—a setting where both precision and recall are critical. VTP generates thin and accurate edge maps that capture both the macro-scale contours of large objects and the micro-scale boundaries of thin structures. The saliency predictions further corroborate this: VTP's maps exhibit tighter, more precisely localized responses on the salient foreground objects, whereas TaskPrompter's saliency bleeds into irrelevant background regions.
 
Across all five tasks, the progressive improvement from VTP\textsubscript{timm} through VTP\textsubscript{vit-5} to VTP\textsubscript{dinov3} is clearly visible, with VFM features amplifying the gains on the most challenging thin structures and tiny objects. This shows that VTP is not merely beneficial for common-case scenarios but is especially advantageous in the long tail of difficult real-world scenes where spatial precision is paramount.

\subsection{Taskonomy Qualitative Results}
 
\begin{figure}[t]
    \centering
    \includegraphics[width=\textwidth]{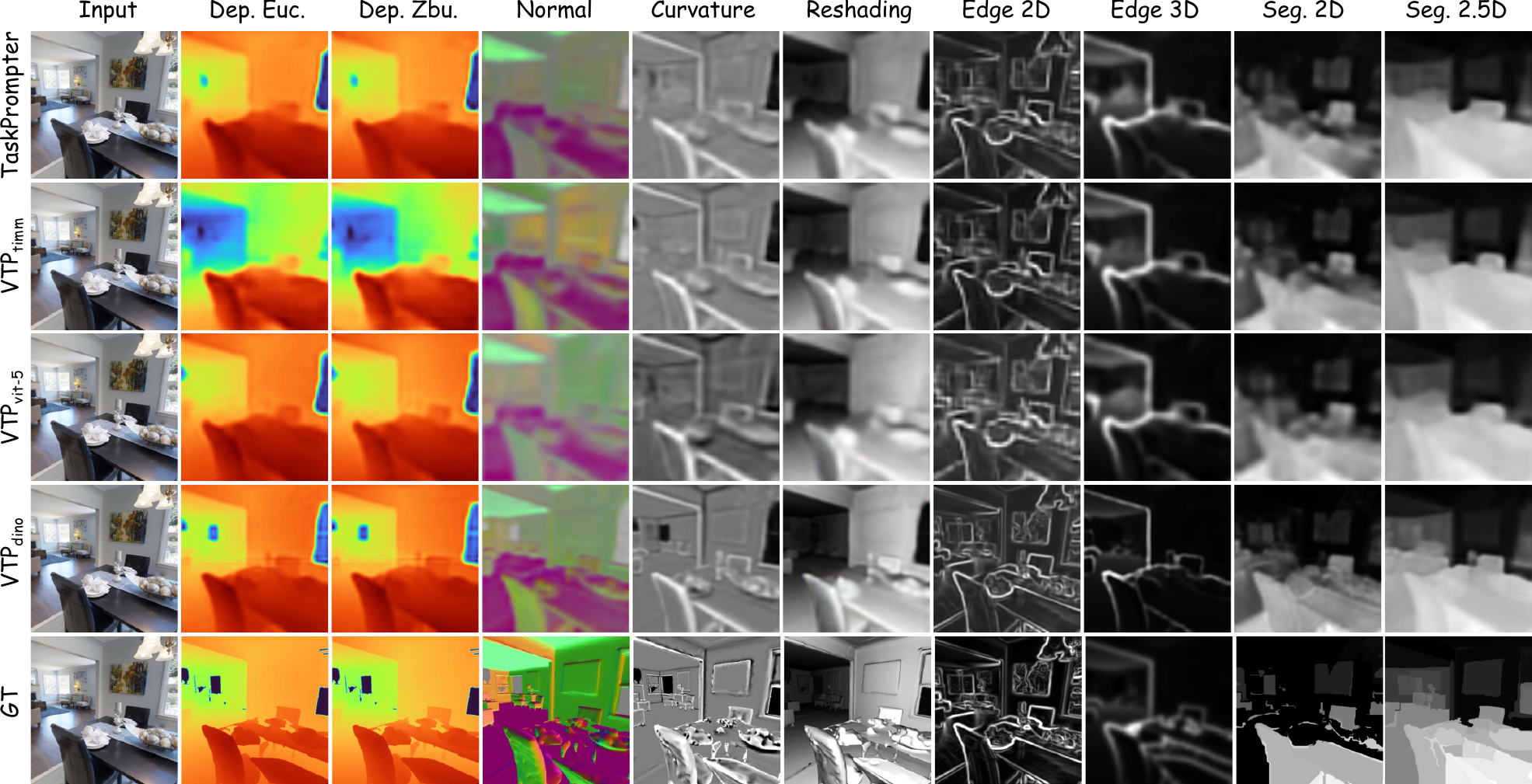}
    \caption{\textbf{Qualitative results on Taskonomy~\cite{zamir2018taskonomy}.} An indoor scene with 9 dense prediction tasks under the extreme 13-task joint training regime. From top to bottom, predictions become progressively finer and more accurate. TaskPrompter~\cite{ye2023taskprompter} produces over-smoothed depth maps, flat curvature, and imprecise edges, while VTP\textsubscript{timm} through VTP\textsubscript{dinov3} successively recover depth discontinuities, surface details, and region coherence, converging toward the ground truth. This progressive improvement requires no additional computational cost. Only the backbone weights are swapped among VFMs.}
    \label{fig:taskonomy_qual}
\end{figure}

Figure~\ref{fig:taskonomy_qual} provides qualitative comparisons on Taskonomy~\cite{zamir2018taskonomy}, which represents one of the most extreme multi-task learning settings in the literature: 13 heterogeneous tasks trained simultaneously, spanning image-level classification, pixel-level geometric estimation, and low-level structural detection. This extreme task diversity creates intense competition for model capacity, making it exceptionally difficult for any single architecture to produce high-quality predictions across all tasks without severe negative transfer. We display 9 of the 13 dense prediction tasks: Euclidean depth (Dep.~Euc.), z-buffer depth (Dep.~Zbu.), surface normals (Normal), curvature (Curv.), reshading (Reshad.), 2D edges (Edge 2D), 3D edges (Edge 3D), 2D segmentation (Seg.~2D), and 2.5D segmentation (Seg.~2.5D). These tasks collectively span the full spectrum from geometric depth and surface estimation to low-level boundary detection and unsupervised region grouping.

The indoor scene in Figure~\ref{fig:taskonomy_qual} features a kitchen workspace with furniture, appliances, and objects at varying depths—a setting that demands accurate geometric reasoning across multiple spatial scales. Reading the figure from top to bottom, a clear trend of progressive improvement is immediately visible: TaskPrompter (top row) produces the coarsest predictions, while VTP\textsubscript{timm}, VTP\textsubscript{vit-5}, and VTP\textsubscript{dinov3} each successively refine the outputs, converging toward the ground truth (bottom row). Crucially, this progressive improvement comes at no extra computational cost or model integration effort as shown in Table~\ref{tab:ablation}, all VTP variants share nearly identical parameters and FLOPs, and the VFM integration requires only swapping the backbone pretrained weights.
 
For the two depth tasks, TaskPrompter tends to produce over-smoothed depth maps that lose the sharp transitions at furniture edges and wall--floor boundaries. The depth discontinuities between the foreground objects and the background are visibly blurred, making it difficult to recover the true 3D layout of the scene. VTP\textsubscript{timm} already preserves these depth boundaries more faithfully. VTP\textsubscript{dinov3} further sharpens the depth predictions, producing maps where the relative ordering and absolute distances of objects are in close agreement with the ground truth for both Euclidean and z-buffer formulations.
 
The curvature and reshading tasks are  revealing, as they require the model to capture subtle surface geometry that is easily overwhelmed in a 13-task setting where gradient signals from diverse tasks compete for shared representation space. TaskPrompter's curvature maps appear flat and lack the fine surface detail visible in the ground truth, while its reshading output misses the nuanced illumination variations on curved surfaces (\textit{e.g.}, the contours of appliances and cabinet edges). VTP progressively recovers these details: VTP\textsubscript{timm} reintroduces coarse surface variation, and VTP\textsubscript{dinov3} further captures the fine-grained curvature and shading gradients that bring the predictions closer to ground-truth. 

For edge detection (both 2D and 3D), the contrast between methods is stark. TaskPrompter's edges are often over-smoothed in heavily textured regions while simultaneously missing true geometric boundaries, resulting in noisy and spatially imprecise edge maps. VTP produces substantially cleaner outputs: the edges are thinner, better localized, and more faithfully aligned with the ground-truth contours. This improvement is particularly pronounced for 3D edges, which rely on geometric rather than photometric cues and thus benefit directly from the richer task-specific representations that VTP's virtual prompts provide.
 
The 2D and 2.5D segmentation tasks complete the picture. TaskPrompter's segmentation maps tend to fragment large planar surfaces into incoherent sub-regions and merge distinct objects that happen to share similar appearance. VTP generates more spatially coherent regions with boundaries that follow the actual object contours, and VTP\textsubscript{dinov3} further improves region purity by leveraging the strong semantic features from DINOv3. The 2.5D segmentation, which incorporates depth cues, shows a consistent advantage for VTP across all variants, confirming that the improved depth representations propagate benefits to downstream tasks that depend on geometric understanding.
 
Taken together, these results demonstrate that VTP's adaptive prompting mechanism scales gracefully to the extreme 13-task regime, consistently improving prediction quality across tasks of vastly different natures—from geometric depth to photometric reshading to structural edge detection—without any further task-specific architectural tuning or additional computational overhead.

\begin{table*}[t]
\centering
\caption{\textbf{Comparisons to further ablating VTP's effectiveness and contribution on Pascal-Context~\cite{chen2014PascalContext}.} 
We compare VTP against (1) standard PEFT (LoRA) on a frozen ViT, (2) HPS with DINOv3 and 
task-specific heads, (3) TaskPrompter forcibly adapted to DINOv3, (4) multi-task 
PEFT methods (Polyhistor, VMT-Adapter, MTLoRA) with their original frozen-backbone 
setting, (5) the same multi-task PEFT methods extended with a fully tunable ViT-L backbone, and 
(6) our VTP$_\text{timm}$. $\Delta_p$ is computed relative to TaskPrompter~\cite{ye2023taskprompter}. The results demonstrate that the effectiveness of VTP is attributed more to its prompting design, rather than the VFMs or the added capacity. Although it shares similar technical formation with those PEFT methods, their goals are different. VTP as a multi-task model architecture jointly trains the base model and the VTP modules, surpassing PEFT methods on both fully and partially fine-tune scenarios.}
\label{tab:appendix_comparison}
\resizebox{\linewidth}{!}{%
\begin{tabular}{@{}l l ccccc c@{}}
\toprule
\multirow{2}{*}{\textbf{Method}} & \multirow{2}{*}{\textbf{Backbone Setup}} & \textbf{Semseg} & \textbf{Parsing} & \textbf{Saliency} & \textbf{Normal} & \textbf{Edge} & {$\Delta_p$} \\
& & mIoU\,$\uparrow$ & mIoU\,$\uparrow$ & maxF\,$\uparrow$ & mErr\,$\downarrow$ & odsF\,$\uparrow$ & (\%)\,$\uparrow$ \\
\midrule
\multicolumn{8}{@{}l}{\cellcolor{gray94}\textit{\textbf{(1) Standard PEFT on Frozen ViT}}} \\
LoRA~\cite{hu2022lora} (r=16) & ViT-L, frozen + LoRA & 74.62 & 63.18 & 82.45 & 14.85 & 71.20 & $-$6.84 \\
LoRA~\cite{hu2022lora} (r=64) & ViT-L, frozen + LoRA & 76.18 & 64.92 & 83.10 & 14.52 & 71.85 & $-$5.21 \\
\midrule
\multicolumn{8}{@{}l}{\cellcolor{gray94}\textit{\textbf{(2) Hard Parameter Sharing with VFM}}} \\
DINOv3~\cite{simeoni2025dinov3} + HPS heads & ViT-L, fully tunable + per-task heads & 81.15 & 70.35 & 84.80 & 13.92 & 73.65 & +0.23 \\
\midrule
\multicolumn{8}{@{}l}{\cellcolor{gray94}\textit{\textbf{(3) TaskPrompter Forcibly Adapted to DINOv3}}} \\
TaskPrompter~\cite{ye2023taskprompter} (DINOv3-init) & ViT-L, weight-mismatched init & 80.45 & 69.74 & 84.78 & 13.82 & 73.65 & +0.27 \\
\midrule
\multicolumn{8}{@{}l}{\cellcolor{gray94}\textit{\textbf{(4) Multi-Task PEFT with Frozen Backbone (original setting)}}} \\
Polyhistor~\cite{liu2022polyhistor} & Swin-T, frozen & 70.87 & 59.54 & 65.47 & 17.47 & 71.85 & $-$8.93 \\
VMT-Adapter~\cite{xin2024vmt} & Swin-T, frozen & 71.60 & 60.67 & 64.02 & 16.41 & 72.10 & $-$8.32 \\
MTLoRA~\cite{agiza2024mtlora} (r=64) & Swin-T, frozen + LoRA & 67.90 & 59.84 & 65.40 & 16.60 & 71.95 & $-$9.21 \\
\midrule
\multicolumn{8}{@{}l}{\cellcolor{gray94}\textit{\textbf{(5) Multi-Task PEFT Extended to Fully Tunable ViT-L Backbone}}} \\
Polyhistor~\cite{liu2022polyhistor} & ViT-L, fully tunable & 77.50 & 65.80 & 83.20 & 14.50 & 72.00 & $-$3.66 \\
VMT-Adapter~\cite{xin2024vmt} & ViT-L, fully tunable & 78.60 & 67.00 & 83.80 & 14.30 & 72.40 & $-$2.50 \\
MTLoRA~\cite{agiza2024mtlora} & ViT-L, fully tunable & 78.10 & 66.40 & 83.50 & 14.40 & 72.20 & $-$3.07 \\
\midrule
\rowcolor{vtpblue!90} \textbf{(6) VTP$_\text{timm}$ (Ours)} & ViT-L, fully tunable & \textbf{82.88} & \textbf{73.85} & \textbf{85.52} & \textbf{13.56} & \textbf{75.12} & \textbf{+2.77} \\
\bottomrule
\end{tabular}%
}
\end{table*}

\section{Visualization of Learned Task Vectors}
\label{sec:app_umap}
 
To gain insights into how task vectors learn and evolve across the model, we analyze the prompt embeddings from two complementary perspectives: (i) activation correlation across tasks and layers, which reveals the pairwise similarity structure of the learned prompts, and (ii) UMAP~\cite{mcinnes2018umap} visualization, which reveals the organization of the prompt embedding space. We conduct this analysis on Pascal-Context with $T = 5$ tasks (semseg, parsing, saliency, surface normals, and boundary detection) and compare TaskPrompter~\cite{ye2023taskprompter} against our VTP.

\subsection{Visualization setup.} For the activation analysis, we compute the pairwise cosine similarity between all task prompt vectors at each ViT block and visualize the resulting $T \times T$ correlation matrices as heatmaps, where red indicates high correlation and white indicates low correlation. This is performed across all $L = 24$ ViT blocks to reveal how inter-task vector relationships develop with network depth. For the UMAP analysis, we extract VTP's $T = 5$ compressed prompt vectors (after the width connection in Eq.~6) at four evenly spaced ViT blocks (blocks 6, 12, 18, and 24), randomly sample 500 images from the Pascal-Context test set, and project the resulting embeddings into a 3D space using UMAP with $n\_\text{neighbors} = 15$, $\text{min\_dist} = 0.1$, and cosine distance. All the points are color-coded by their task identities and are annotated by layer.

\subsection{TaskPrompter: anomalous activation patterns.} As shown in Figure~\ref{fig:umap}~(a), TaskPrompter's task vectors exhibit progressive over-alignment with network 
depth. At shallow layers, the correlation matrix shows moderate inter-task 
similarity with discernible pair-wise structure, indicating that the task 
vectors initially encode differentiated information. However, as depth 
increases, this structure is rapidly eroded: by layer 18, the mean 
off-diagonal correlation reaches $0.87$ with a min–max range of only $0.04$, 
and by layer 24 it further rises to 0.94 with a comparably narrow range. 
This pattern reveals that the issue is not merely increasing similarity, 
but the loss of task-pair specificity—all pairs converge toward the same 
high-correlation value regardless of their semantic relationship. Such 
uniform saturation indicates representational collapse, where the task 
vectors no longer encode meaningful task-specific distinctions and instead 
degenerate into near-identical directions in the embedding space.

\subsection{VTP: well-structured UMAP clusters.} Figure~\ref{fig:umap}~(b) presents the 3D UMAP projection of VTP's compressed prompt embeddings across four representative layers. It shows clearly separated clusters, with each task forming a distinct, compact group in the embedding space. The five task clusters are well-separated from one another, indicating that VTP's read-write cycle successfully distills task-discriminative information from the richer $T + V$ virtual prompt bank into the $T$ prompts that interact with image tokens. This structure confirms that the read-write routing matrices learn to organize the prompt space in a way that preserves both task specificity and cross-task sharing, which is precisely the balance that effective multi-task learning requires. Furthermore, the layer-wise information encoded in the UMAP reveals that task vectors within each layer form an internally cohesive latent subspace tailored to that layer's specific representational role, rather than collapsing into a single shared representation.

\subsection{VTP: progressive activation specialization.} Figure~\ref{fig:umap}~(c) shows VTP's activation correlation matrices across the same 24 ViT blocks, providing a direct counterpart to the TaskPrompter analysis in (a). Holistically, VTP preserves a layer-dependent task-affinity geometry. In 
early layers, the strongest correlations appear between tasks that share 
fine-grained geometric and boundary cues, particularly surface normals and 
edges. As depth increases, semantic pairs such as semantic segmentation and 
saliency become increasingly aligned, reflecting the transition from local 
boundary processing to object-level reasoning. Importantly, this evolution 
does not collapse into uniform high correlation: weakly related task pairs 
remain clearly separated from strongly related ones throughout all layers. 
This structured evolution contrasts sharply with TaskPrompter, whose 
deep-layer task vectors become nearly homogeneous across all task pairs 
(off-diagonal range of $0.05$ at layer 24, compared to $0.29$ for VTP).

\section{Discussion on Flexibility and VFM Compatibility}
\label{sec:app_flexibility}
 
In the main text, we describe VTP as a flexible, drop-in prompting architecture that preserves full compatibility with off-the-shelf VFMs. Below, we discuss these properties and compare them with existing multi-task scene understanding architectures, and outline the integration procedure of VFMs.
 
\subsection{Drop-in Compatibility}
 
The aforementioned drop-in compatibility indicates that VTP can be attached to existing standard pretrained ViTs without modifying its internal block topology, weight shapes, or forward-pass logic.

\vspace{-0.60em}
\paragraph{Unmodified ViT block internals.} 
Multi-head self-attention, feed-forward networks (FFNs), layer normalization, and residual connections within each standard ViT block remain untouched. Rather than inserting specifically tailored attention heads or altering the residual streams, VTP simply extends the input image sequence by $T$ task tokens along the sequence dimension, which is an operation natively supported by standard Transformer implementations~\cite{Oquab2023DINOv2, simeoni2025dinov3}.

\vspace{-0.60em}
\paragraph{Exact weight-shape preservation.} 
Since VTP preserves the ViT block's internal structure, all pretrained weight tensors retain their original pretrained shapes and can be loaded without any remapping, pruning, or interpolation. This is in direct contrast to previous multi-task prompting methods like TaskPrompter~\cite{ye2023taskprompter}, which modifies the inner self-attention module to incorporate spatial-channel dual attention for enhanced representation capacity. Such modifications inevitably alter the weight shapes of the attention projection matrices, making it difficult to directly load off-the-shelf VFM weights without re-engineering the model architectures. VTP avoids this issue entirely: the pretrained backbone weights are loaded as-is, and only the VTP-specific parameters (prompt embeddings, read-write routing matrices, prediction heads) are initialized from scratch.

\vspace{-0.60em}
\paragraph{Preserved macro architecture.} 
VTP does not alter the number of ViT blocks, the embedding dimension, the number of attention heads, or the feed-forward expansion ratio. Operating entirely as an external bypass that interfaces with the encoder at block boundaries, VTP supports different ViT scales (e.g., ViT-B, ViT-L, ViT-H) without further structural adaptation.

\vspace{-0.60em}
\paragraph{Compatibility across VFM families.} 
We validate this property empirically by deploying VTP on three distinct weight sources: (i) standard ImageNet-supervised pretraining (ViT-L from \texttt{timm}~\cite{dosovitskiy2020image}), (ii) DINOv3~\cite{simeoni2025dinov3}, a self-supervised VFM trained with a distillation-based objective, and (iii) ViT-5~\cite{wang2026vit5}, a recent VFM trained with a hybrid supervised--self-supervised pipeline. These three represent qualitatively different pretraining paradigms, yet VTP integrates with each by simply swapping the backbone weights—no code changes, no hyperparameter re-tuning for the VTP itself. The results in Tables~1 and~2 confirm that VTP benefits from each VFM's strengths without compatibility issues.

\subsection{Architectural Flexibility}

Beyond VFM compatibility, VTP offers flexibility across three practical dimensions:

\vspace{-0.60em}
\paragraph{Flexibility in model choice.} As aforementioned, VTP is agnostic to the specific ViT variant and pretrained weights used as the backbone. This allows practitioners to adopt their available VFM for their domain, \textit{e.g.}, a medical imaging VFM for clinical applications, a remote sensing VFM for satellite imagery, without any modifications to the VTP module. The only need is that the backbone follows the standard ViT block structure: multi-head self-attention followed by a feed-forward network, with layer normalization and residual connections.

\vspace{-0.60em}
\paragraph{Task scalability.}
VTP accommodates varying numbers of tasks without structural changes to the architecture. Adding a new task requires only: (a) adding one new learnable prompt embedding (increasing $T$ by 1), (b) expanding the read-write routing matrices by one row/column, and (c) adding a new lightweight prediction head. The encoder backbone and the VTP module's core logic remain unchanged. This is validated empirically by our experiments spanning 4 tasks (NYUD-v2), 5 tasks (Pascal-Context), and 13 tasks (Taskonomy), all using the same VTP architecture with only the task count adjusted. The virtual task number $V$ can remain fixed across these settings as shown in Table~\ref{tab:ablation}.

\vspace{-0.60em}
\paragraph{Flexibility in deployment.} Since VTP's external routing operates solely on the $T{+}V$ task stream, it introduces minimal parameter and FLOP overhead when $T, V \ll N$. As shown in Table~\ref{tab:ablation} (right), VTP$_\text{timm}$ reduces parameter count (342M vs.\ 401M) and increases inference throughput (16.3 vs.\ 14.4 imgs/s) compared to TaskPrompter~\cite{ye2023taskprompter}, making it well-suited for resource-constrained deployments.

\subsection{Compatibility Comparison with Prior MTL Architectures}

Existing multi-task scene understanding architectures face distinct barriers when adapting off-the-shelf VFMs: 
\textbf{(i) TaskPrompter~\cite{ye2023taskprompter}} replaces standard self-attention with spatial-channel dual attention and channel-wise sliding windows, creating weight-shape mismatches that preclude direct checkpoint loading. 
\textbf{(ii) InvPT~\cite{ye2022invpt} and InvPT++~\cite{ye2024invplpl}} interleave cross-attention modules that tightly couple multi-scale decoder projections to backbone feature distributions, requiring careful per-VFM recalibration. 
\textbf{(iii) TaskExpert~\cite{ye2023taskexpert}} inserts dynamic mixture-of-experts gating between encoder layers, similarly coupling routing calibration to specific backbone hidden states. 
\textbf{(iv) MTMamba~\cite{lin2024mtmamba} and PAME~\cite{yu2026pame}} rely on state-space decoders with specialized scanning patterns tailored to specific feature hierarchies, which complicates backbone substitution.



\end{document}